\documentclass{article}

\usepackage{arxiv}

\usepackage[utf8]{inputenc} % allow utf-8 input
\usepackage[T1]{fontenc}    % use 8-bit T1 fonts
\usepackage{hyperref}       % hyperlinks
\usepackage{url}            % simple URL typesetting
\usepackage{booktabs}       % professional-quality tables
\usepackage{amsfonts}       % blackboard math symbols
\usepackage{nicefrac}       % compact symbols for 1/2, etc.
\usepackage{microtype}      % microtypography
\usepackage{lipsum}
\usepackage{graphicx}
\usepackage{cite}
\usepackage{comment}
\usepackage{algorithmic}
\usepackage{array}
\usepackage{eqparbox}
\usepackage[tight,footnotesize]{subfigure}
\usepackage{multicol}
\usepackage{url}
\usepackage{tabularx,ragged2e,booktabs,caption}
\usepackage{subfigure} 
\usepackage{multicol}
\usepackage{footmisc}[bottom]
\usepackage[export]{adjustbox}
\usepackage{algorithm}
\usepackage{algorithmic}
\usepackage{enumerate}
\usepackage{hyperref,mathtools}
\usepackage{breakcites}
\usepackage{blindtext}
\usepackage{booktabs}
\usepackage{amssymb}
\usepackage{lipsum}

\title{Predicting Shot Making in Basketball Learnt from Adversarial Multiagent Trajectories}

\author{
 Mark Harmon \\
  Northwestern University\\
  \texttt{mdharmo13@gmail.com} \\
   \And
 Abdolghani Ebrahimi \\
  Northwestern University\\
  \texttt{ghani@u.northwestern.edu} \\
  \And
 Patrick Lucey \\
  Stats Perform\\
  \texttt{patrick.lucey@statsperform.com} \\
  \And
 Diego Klabjan \\
  Northwestern University\\
  \texttt{d-klabjan@northwestern.edu} \\
}
\begin{document}
\maketitle
\begin{abstract}
In this paper, we predict the likelihood of a player making a shot in basketball from multiagent trajectories.  Previous approaches to similar problems center on hand-crafting features to capture domain specific knowledge.  Although intuitive, recent work in deep learning has shown this approach is prone to missing important predictive features.  To circumvent this issue, we present a convolutional neural network (CNN) approach where we initially represent the multiagent behavior as an image.  To encode the adversarial nature of basketball, we use a multichannel image which we then feed into a CNN.  Additionally, to capture the temporal aspect of the trajectories we use ``fading.''  We find that this approach is superior to a traditional FFN model.  By using gradient ascent, we were able to discover what the CNN filters look for during training.  Last, we find that a combined FFN+CNN is the best performing network with an error rate of 39\%.
\end{abstract}

% keywords can be removed
%\keywords{First keyword \and Second keyword \and More}

\section{Introduction}
Neural networks have been successfully implemented in a plethora of prediction tasks ranging from speech interpretation to facial recognition. Because of ground-breaking work in optimization techniques (such as batch normalization \cite{ioffe2015batch} ) and model architecture (convolutional, deep belief, and LSTM networks), it is now tractable to use deep neural networks to effectively learn a better feature representation compared to hand-crafted methods. 

One area where such methods have not been utilized is the space of adversarial multiagent systems (for example, multiple independent players in competition), specifically when the multiagent behavior comes in the form of trajectories. There are two reasons for this: i) procuring large volumes of data where deep methods are effective is difficult to obtain, and ii) forming an initial representation of the raw trajectories so that deep neural networks are effective is challenging. In this paper, we explore the effectiveness of deep neural networks on a large volume of basketball tracking data, which contains the $x, y$ locations of multiple agents (players) in an adversarial domain (game).

To thoroughly explore this problem, we focus on the following task: ``given the trajectories of the players and ball in the previous five seconds, can we accurately predict the likelihood that a player with position/role $X$ will make the shot?" For this paper, player role refers to a more fluid position of a player, which was explored by \cite{lucey2013representing}.  For example, a player may not be in the point guard position during the entire play. Since we plan to utilize an image representation for player trajectories, we use a convolutional neural network (CNN), which is widely considered a powerful method for image classification problems. 

In this study, we treat each player as a generic player, i.e. we are not using player identities. By modeling generic players rather than individuals, we can more easily quantify the differences between an individual player and a generic player. Consider, for example, a rookie (which clearly has limited historical data). After a few games, a coach can compare his shooting performance against the performance of a generic player, which can be obtained by our model. Our model offers such comparisons at a very fine granular level of play. The coach can then guide the player to improvements. The same argument is applicable in other cases such as a player getting more play-time (the model allows comparison against generic players in addition to his past performance, which clearly does not need our model). To obtain these values, we want to identify shooting for each generic position (point guard, shooting guard, center, power forward, and small forward). Since we classify whether the shot will be made for a single offensive position, every offensive player corresponds to either the class of making a shot or missing a shot. Therefore, our classification problem consists of ten classes.

Our work contains three main contributions. First, we represent trajectories for the offense, ball, and defense as an eleven channel image. Each channel corresponds to the five offensive and defensive players, as well as the ball. To encode the direction of the trajectories, we fade the paths of the ball and players. In our case, an instance is a possession that results in a shot attempt. Second, we apply a combined convolutional neural network (CNN) and feed forward network (FFN) model on an adversarial multiagent trajectory based prediction problem. Third, we gain insight into the nature of shot positions and the importance of certain features in predicting whether a shot will result in a basket.

Our results show that it is possible to solve this problem with relative significance. The best performing model, the CNN+FFN model, obtains an error rate of 39\%. In addition, it can accurately create heat maps by shot location for each player role. Features which are un-surprisingly important are the number of defenders around the shooter and location of the ball at the time of the shot. In addition, we found that our image-based model performs just as well as a feed-forward technique that required us to build nearly 200 features.

\section{Literature Review}

With the rise of deep neural networks, sports prediction experts have new tools for analyzing players, match-ups, and team strategy in these adversarial multiagent systems. Trajectory data was not available at the time, so much previous work on basketball data utilizing the power of neural networks have used statistical features such as: the number of games won and the number of points scored. For example, \cite{loeffelholz2009predicting} use statistics from 620 NBA games and a neural network to predict the winner of a game. Another work in predicting game outcomes is \cite{mccabe2008artificial}. On the other hand, \cite{nalisblog} in his blog discusses predicting basketball shots based upon the type of shot (layups versus free throws and three-point shots) and where the ball was shot from. In a recent paper, \cite{murakamianalysis} used 12 important features in predicting the shot success using neural networks. In other sports related papers, \cite{huang2010neural} use a neural network to predict winners of soccer games in the 2006 World Cup. Also, \cite{wickramaratna2005neural} predict goal events in video footage of soccer games. 

With trajectory data becoming available, some more recent works in the area utilize that in their basketball game analysis. For instance, \cite{lucey2014get} uses the data set provided by STATS LLC to explore how to get an open shot in basketball using trajectory data to find that the number of times defensive players swapped roles/positions was predictive of scoring. However, they explore open versus pressured shots (rather than shot making prediction), do not represent the data as an image, and do not implement neural networks for their findings. Other trajectory work includes using Conditional Random Fields to predict ball ownership from only player positions \cite{wei2015predicting}, as well as predicting the next action of the ball owner via pass, shot, or dribble \cite{yue2014learning}.  Non-negative matrix factorization is used by \cite{miller2014factorized} to identify different types of shooters using trajectory data. Because of a lack of defensive statistics in the sport, studies \cite{franks2015characterizing, franks2015counterpoints} create counterpoints (defensive points) to better quantify\, defensive plays. The work by \cite{pervse2009trajectory} makes use of trajectory data by segmenting a game of basketball into phases (offense, defense, and time-outs) to then analyze team behavior during these phases. 

The work \cite{wang2016classifying} use trajectory data representations and recurrent neural networks (rather than CNN's) to predict plays. Because of the nature of our problem, predicting shot making at the time of the shot, there is not an obvious choice of labeling to use for a recurrent network. They also fade the trajectories as the players move through time. Like us, they create images of the trajectory data of the players on the court. Our images differ in that we train our network on the image of a five second play and entire possession, while their training set is based on individual frames represented as individual positions rather than full trajectories. They use the standard RGB channels, which we found is not as effective as mapping eleven channels to player roles and the ball for our proposed classification problem. Also, the images they create solely concentrate on the offense and do not include defensive positions. 

In addition, work by \cite{cervone2016multiresolution} explores shots made in basketball using the same STATS data using a multiresolution stochastic process. Our work focuses on a representation capturing the average adversarial multiagent behavior, compared to the approach of \cite{cervone2016multiresolution}, which focuses on individual players. In addition, while they concentrate on a Markov model that transitions between a coarse and fine data representation, our goal is to represent the fine-grained data for a deep learning model. Our work focuses on shot prediction for the average player in the NBA at the time of the shot while their work formulates a way to calculate estimated point value based upon specific player identity without reporting predictions for individual plays. Their estimated point value includes the probability of making a shot; however, this probability is based on individual characteristics of a player prior to taking the shot which is different from our goal of considering generic players. We do not see an easy way to modify their models to provide answers based on our setting. In summary, there is no readily available numerical comparison of the two models that would provide answers for shot making predictions of generic players.

The final model that we implement, the combined network, utilizes both image and other statistical features. There is work that utilizes both image and text data with a combined model. Recently, \cite{xu2015show, karpathy2015deep, socher2014grounded, vinyals2015show, mao2014deep}, all explore the idea of captioning images, which requires the use of generative models for text and a model for recognizing images. However, to the best of our knowledge, we have not seen visual data that incorporates fading an entire trajectory for use in a CNN.

\section{Data}

The dataset was collected via SportsVU by STATS LLC. SportsVU is a tracking system that uses 6 cameras to track all player locations (including referees and the ball). The data used in this study was from the 2012-2013 NBA season and includes thirteen teams, which have approximately forty games each. Each game consists of at least four quarters, with a few containing overtime periods. 

The SportVU system records the positions of the players, ball, and referees 25 times per second. At each recorded frame, the data contains the game time, the absolute time, player and team anonymized identification numbers, the location of all players given as (x, y) coordinates, the role of the player, and some event data (i.e. passes, shots made/missed, etc.). It also contains referee positions, which are unimportant for this study, and the three-dimensional ball location. Table \ref{dataexampbball} shows a sample (with player identities masked) of a single frame of data.

\begin{table}
\centering
\caption{Trajectory Data Sample} \label{tab:title}
\resizebox{\columnwidth}{!}{
\begin{tabular}{rrrrrrrrr}
\hline
Game Time& Real Time & Team & Player & X & Y & Z & Role & Event \\
\hline
693 & 514200& 1 & 101 & 21.5 & 33.6 & 0.0 & 1 & 0 \\
693 & 514200& 1 & 102 & 24.1 & 14.1 & 0.0 & 2 & 1 \\
693 & 514200& 1 & 103 & 5.4 & 9.6 & 0.0 & 3 & 0 \\
693 & 514200& 1 & 104 & 3.9 & 45.6 & 0.0 & 4 & 0 \\
693 & 514200& 1 & 105 & 10.4 & 3.5 & 0.0 & 5 & 0 \\
693 & 514200& 2 & 201 & 13.6 & 31.6 & 0.0 & 1 & 0 \\
693 & 514200& 2 & 202 & 20.4 & 15.4 & 0.0 & 2 & 0 \\
693 & 514200& 2 & 203 & 7.7 & 13.3 & 0.0 & 3 & 0 \\
693 & 514200& 2 & 204 & 6.0 & 38.6 & 0.0 & 4 & 0 \\
693 & 514200& 2 & 205 & 13.9 & 13.2 & 0.0 & 5 & 0 \\
693 & 514200& -1 & -1 & 25.1 & 14.0 & 3.4 & 0 & 0 \\
693 & 514200& -2 & 1 & 16.9 & 49.2 & 0.0 & 0 & 0 \\
693 & 514200& -2 & 3 & 78.9 & 0.5 & 0.0 & 0 & 0 \\
693 & 514200& -2 & 2 & 26.0 & 3.1 & 0.0 & 0 & 0 \\
\hline
\end{tabular}
}
\label{dataexampbball}
\end{table}

The sample detailed above shows a single frame snapshot from a game where team 1 is playing team 2. "Game Time" refers to time left in the quarter in seconds while "Real Time" is the actual time of the day outside of the game. The next column is the team labels where the ball and referees are denoted with a "-1" and "-2," respectively. Each player has an ID along with the ball and each referee. Next are the coordinates of all players, referees, and the ball along with the role/position of the player and special event codes for passes, shots, fouls, and rebounds.

This dataset is unique in that before SportVU, there was very little data available of player movements on the court and none known that provides frame-by-frame player locations. Since it is likely that most events in basketball can be determined by the movements of the players and the ball, having the trajectory data along with the event data should provide a powerful mixture of data types for prediction tasks.

There are a few ways to extract typical shot plays from the raw data. One is to choose a flat amount of time for a shooting possession. In our case we choose to include the final five seconds of a typical possession. To obtain clean plays, those that lasted less than 5 seconds due to possession changes and those in which multiple shots were taken were thrown out. After throwing out these cases, we were left with 75,000 five second plays.

The other way of obtaining play data would be to take the entire possession. Thus, rather than having plays be limited to five seconds, possessions can be much longer or shorter. Since the raw data does not contain labels for possession, we had to do this ourselves. To identify possession, we calculate the distance between the ball and each of the players. The player closest to the ball would be deemed the ball possessor. Since this approach may break during passes and other events during a game, we end possession when the ball is closer to the defensive team for 12 frames (roughly 0.5 seconds). The procedure yields 72,000 possession examples. We found that using an entire possession resulted in lower prediction accuracy probably due to additional intricacies of court positions. Therefore, we use five second possessions for the remainder of this study.

Although each player has an assigned role, players change positions with respect to each other, which effectively changes their role during regular play \cite{lucey2013representing}. Since our classification problem is dependent upon a player's position, a player's role must be chosen for each five second play. We ultimately decide to assign a player the role that they occupy at the beginning of the play. Since we want to explore how a player's actions resulted in favorable/unfavorable shooting position, we do not assign role based upon the end of the five second play.

\section{Image-Based Representation}

\begin{figure}[ht]
\centering
\begin{minipage}[b]{0.2\textwidth}
\includegraphics[width=\textwidth]{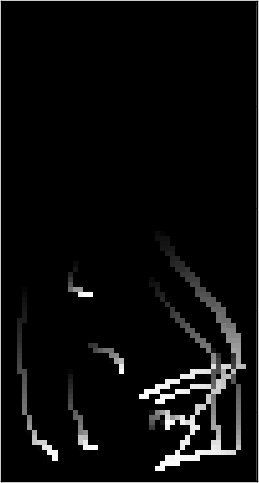}
\end{minipage}
\begin{minipage}[b]{0.2\textwidth}
\includegraphics[width=\textwidth]{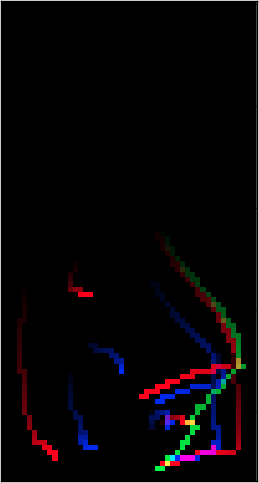}
\end{minipage}
\caption{A grayscale and RGB image of the same trajectory of all 10 players plus the ball. Each red/blue line corresponds to an offensive/defensive player, and green indicates the ball trajectory.}
\label{cnnes}
\end{figure}

In terms of applying deep neural networks to multiagent trajectories, we first form an initial representation. A natural choice for representing these trajectories is in the form of an image. Given that the basketball court is $50$x$94$ feet, we can form a $50$x$94$ pixel image. In terms of the type of image we use, there are multiple choices: i) grayscale (where we identify that a player was a specific location by making that pixel location 1), ii) RGB (we can represent the offense trajectories in the red channel, the defense in the blue channel and the ball in the green channel, and the occurrence of a player/ball at that pixel location can be representing by a 1), and iii) 11-channel image (where each agent has their own separate channel) with each position represented by a 1. Examples of the grayscale and RGB approach are shown in Figure \ref{cnnes}. 

The 11-channel approach requires some type of alignment. In this paper, we apply the `role-representation' which was first deployed by \cite{lucey2013representing}. The intuition behind this approach is that for each trajectory, the role of that player is known (i.e., point-guard, shooting guard, center, power-forward, small-forward). This is found by aligning to a pre-defined template which is learnt in the training phase. 

Figure \ref{cnnes} shows examples of the methods we use to represent our data for our CNN. The grayscale image, which appears on the left, can accurately depict the various trajectories in our system. However, because the image is grayscale, the CNN will treat each trajectory the same. Since we have an adversarial multiagent system in which defensive and offensive behavior in trajectory data can lead to different conclusions, grayscale is not the best option for representation. Therefore, to increase the distinction between our agents, we represent the trajectories with an RGB scale. We choose red to be offense, blue to be defense, and green to be the ball. This approach takes advantage of multiple channels to allow the model to better distinguish our adversarial agents. Although the ball may be part of the offensive agent structure, we decide to place the ball in a channel by itself since the ball is the most important agent. This approach, although better than the gray images, lacks in distinguishing player roles. Since we classify our made and missed shots along with the role of the player that shoots the ball, a CNN will have trouble distinguishing the different roles on the court. Therefore, for our final representation, we decide to separate all agents into their own channel so that each role is properly distinguished by their own channel.

\begin{figure}[ht]
\centering
\begin{minipage}[b]{0.2\textwidth}
\includegraphics[width=\textwidth]{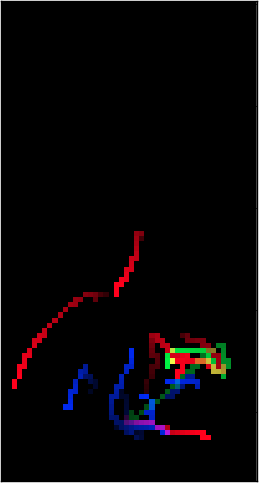}
\end{minipage}
\begin{minipage}[b]{0.2\textwidth}
\includegraphics[width=\textwidth]{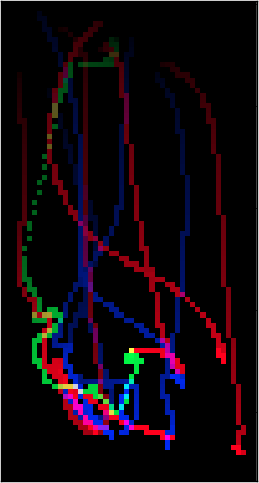}
\end{minipage}
\caption{Figure on left depicts a five second play with all 10 players while the right is of an entire possession. As expected of five second plays, most of the trajectories remain near the basket located at the bottom of the image. Each red/blue line corresponds to a offensive/defensive player with green indicating the ball trajectory.}
\label{traj}
\end{figure}

The above ideas nearly create ideal images; however, it does not include time during a play. Since each trajectory is of equal brightness from beginning to end, it may be difficult to identify where the ball was shot from and player locations at the end of the play. Therefore, we implement a fading variable at each time frame. We subtract a parameterized amount from each channel of the image to create a faded image as seen in Figures \ref{cnnes} and \ref{traj}. Thus, it becomes trivial to distinguish the end of the possession from the beginning and leads to better model performance.
%\begin{figure*}[ht]
%\centerline{\subfloat[Case I]\includegraphics[width=2.5in]{subfigcase1}%
%\label{fig_first_case}}
%\hfil
%\subfloat[Case II]{\includegraphics[width=2.5in]{subfigcase2}%
%\label{fig_second_case}}}
%\caption{Simulation results}
%\label{fig_sim}
%\end{figure*}

\section{Models}

To fully utilize the power of this dataset, we implement a variety of networks for our prediction task. For our base model, we use logistic regression with 197 hand-crafted features detailed later. To improve upon this basic model, we use a multilayer FFN with the same features and utilize batch normalization during training. Because of the nature of these two models, we could only include the positions of the players at the time of the shot. Therefore, we craft images to include the position of the players throughout the possession. We then apply a CNN to these new image features. Finally, we create a combined model that adopts both images and the original FFN features for training.

\subsection{Logistic Regression and Feed Forward Network}

For the baseline models, features based upon basketball knowledge were crafted. The list of features includes:

\begin{itemize} 
\item Player and ball positions at the time of the shot 
\item Game time and quarter time left on the clock
\item Player speeds over five seconds
\item Speed of the ball
\item Distances and angles (with respect to the hoop) between players
\item Number of defenders in front of the shooter ($30^{0}$ angle of the shooter) and within six feet based upon the angles calculated between players
\item Ball possession time for each offensive player
\item Number of all individuals near the shooter (including teammates)

\end{itemize}

Logistic regression and FFN both use the same calculated features. In addition, only the CNN does not incorporate the above features.

A deep neural network is a machine learning model that consists of several layers of linear combinations of weights, inputs, and activation functions. The number of weights $\theta$, layers $L$, and activation functions $a$ are specified before training with data $X$. The model outputs probabilities $f$ and the error is calculated generally with the Kullback-Leibler divergence, a.k.a log loss function $KL(y||f)$, against the true values $y$. A deep neural network generally refers to a neural network that is at least three layers deep. The depth (number of layers) of network versus the breadth (number of neurons)  allows it to learn more complex features of the dataset. The following is a mathematical model of a deep neural network:

\begin{align*}
&z^{0} = X  \enspace \text{inputs of the model,}\\
&z_{i}^{\ell} (X;\theta) = a^{\ell} \Big( \sum_{k=1}^{K} z_{k}^{\ell -1} \theta_{ik}^{\ell - 1}\Big) \enspace\text{for neuron $i$ in layer $\ell$}\\ &\text{and $K$ neurons at layer $\ell - 1$,}\\
&f_i (X;\theta) = \frac{e^{z_{i}^{L} {\theta_{i}^{L}}^{T}}}{\sum_m e^{z_{m}^{L} {\theta_{i}^{L}}^{T}}} \enspace  \text{softmax function for probability}\\ &\text{of each class $i$,}
\end{align*}
with the optimization function:
\begin{equation}
    \min_\theta E_{(X,Y)} KL(Y||f(X,;\theta)) = E_{(X,Y)} \sum_{i} y_i \log f_i (X;\theta)
\end{equation}

\subsection{Convolutional Neural Network}

A CNN is similar to a Feed Forward Network, except that instead of learning individual weights per neuron, the model learns many filters (which consist of weights) that are convolved with the incoming data and reduced in size by a pooling layer. Like FFN's, they consist of multiple layers. For our model, we use a CNN that consists of three full convolutional layers, each with 32 3x3 filters and a max-pooling layer with a pool-size of 2x2 following each convolutional layer. After the final pooling layer, there is a fully connected layer with 400 neurons. The network ends with an output layer consisting of a softmax function and ten neurons. In addition, we use the ReLU function for our nonlinearity at each convolutional layer and the fully connected layer. We also implement AlexNet, Network-in-Network, and Residual Networks, but we did not garner significant improvement from any of these models.

\subsection{CNN + FFN Network}

The final network implemented is a combination of both the feed forward and convolutional networks. For this model, we use both the feed forward features and the fading trajectory images from the CNN. The idea behind the combined network is to have the model identify trajectory patterns in the images along with statistics that are known to be important for a typical basketball game. 

The CNN and FFN parts of the combined network have the exact same architecture as the stand-alone versions of each model. The final layers just before the softmax layer of each stand-alone network are then fully-connected to a feed-forward layer that consists of 1,000 neurons. This layer is then fed into the final softmax layer to give predictions. After performing experiments and measuring log loss, we found that adding layers to this final network or adding additional neurons to this layer did not improve our final results.

\section{Results}

All the models (FFN, CNN, and CNN+FFN) use the typical log loss function as the cost function with a softmax function at the output layer. The weights are initialized with a general rule of $\pm\sqrt{1/n}$ where $n$ is the number of incoming units into the layer. For training, we implement the batch stochastic gradient method utilizing batch normalization. Batch normalization is used on the convolutional and feed forward layers of the model. In addition, we train the models on a NVIDIA Titan X using Theano. 

For the CNN and CNN+FFN networks we utilize our eleven channel images with fading. We randomly split our data into train, validation, and test sets. The training set contains 52,000 samples while the validation and test sets contain 10,000 samples each. All experiments are completed using the same data split.

\begin{figure}[ht]
\includegraphics[scale=0.45]{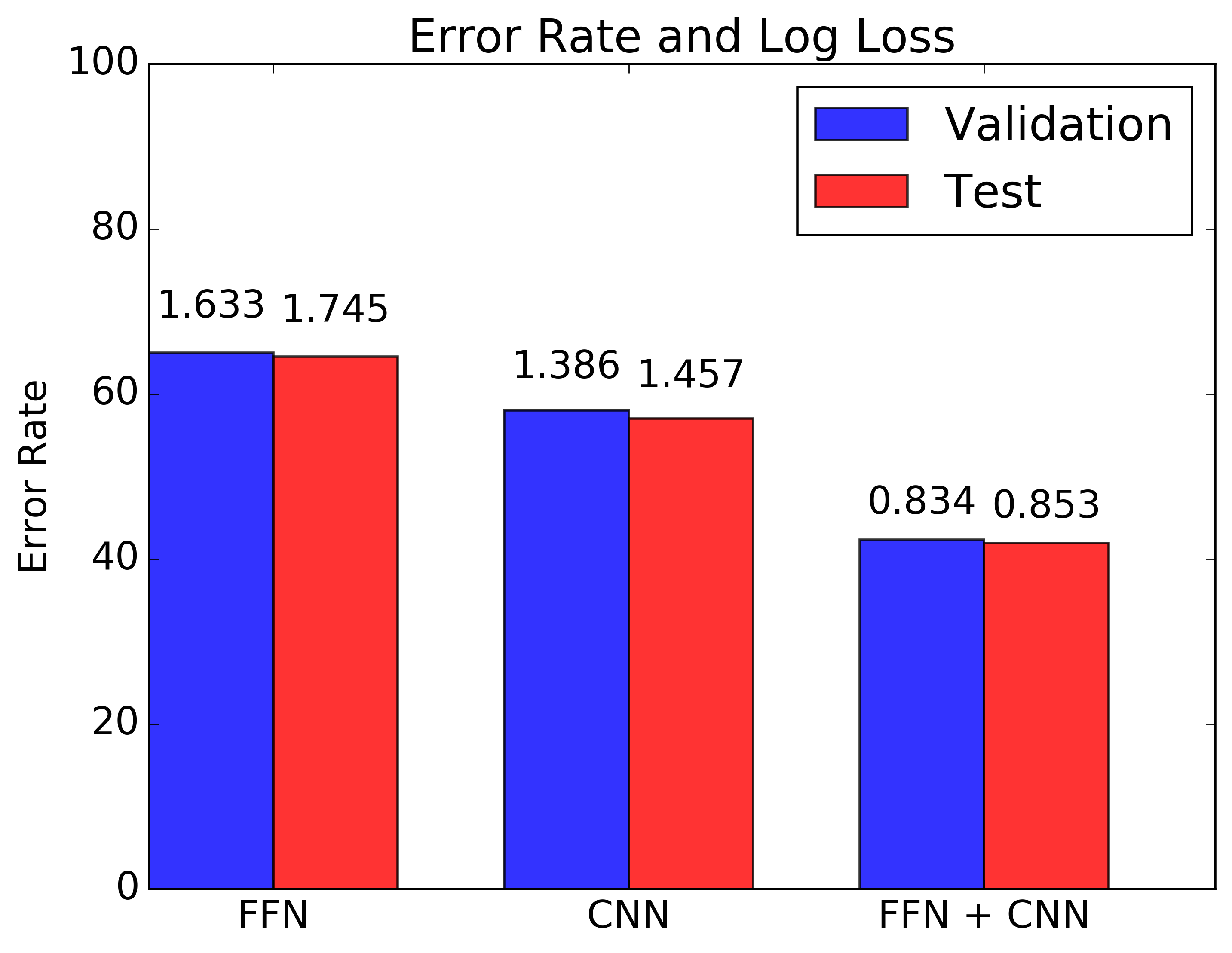}
\centering
\caption{Log loss and accuracy for three image representations (smaller is better). The eleven channel method is superior.}
\label{rep}
\end{figure}

To justify our image representation, we first evaluate our model with each of the following image sets: one (grey) channel, three (RGB) channels, and eleven channels (for each player and the ball). These sets are all assessed on the previously mentioned CNN architecture consisting of three convolutional layers. Figure \ref{rep} displays the log loss and error rate on both validation and test sets for each of our representations. The log loss and error rate show a dramatic difference in accuracy based upon each image representation. By both accuracy and loss metrics, using eleven channels is the best representation. The eleven channel representation minimizes the overlapping trajectory issue that hinders the one and three channel methods. Thus, with eleven channels, the CNN can more easily capture more relevant on-the-court features such as ball possession and passes. 

In addition to the image representations of Figure \ref{traj}, we made each trajectory a different color in RGB space, varying the strength of the fading effect, and including extra channels of heat maps depicting the final ball position (none of which outperformed the eleven channel method). Successfully representing trajectory data in sport that outperforms traditional metrics is a nontrivial problem, but is not further explored in this study.

\begin{figure}[ht]
\includegraphics[scale=0.45]{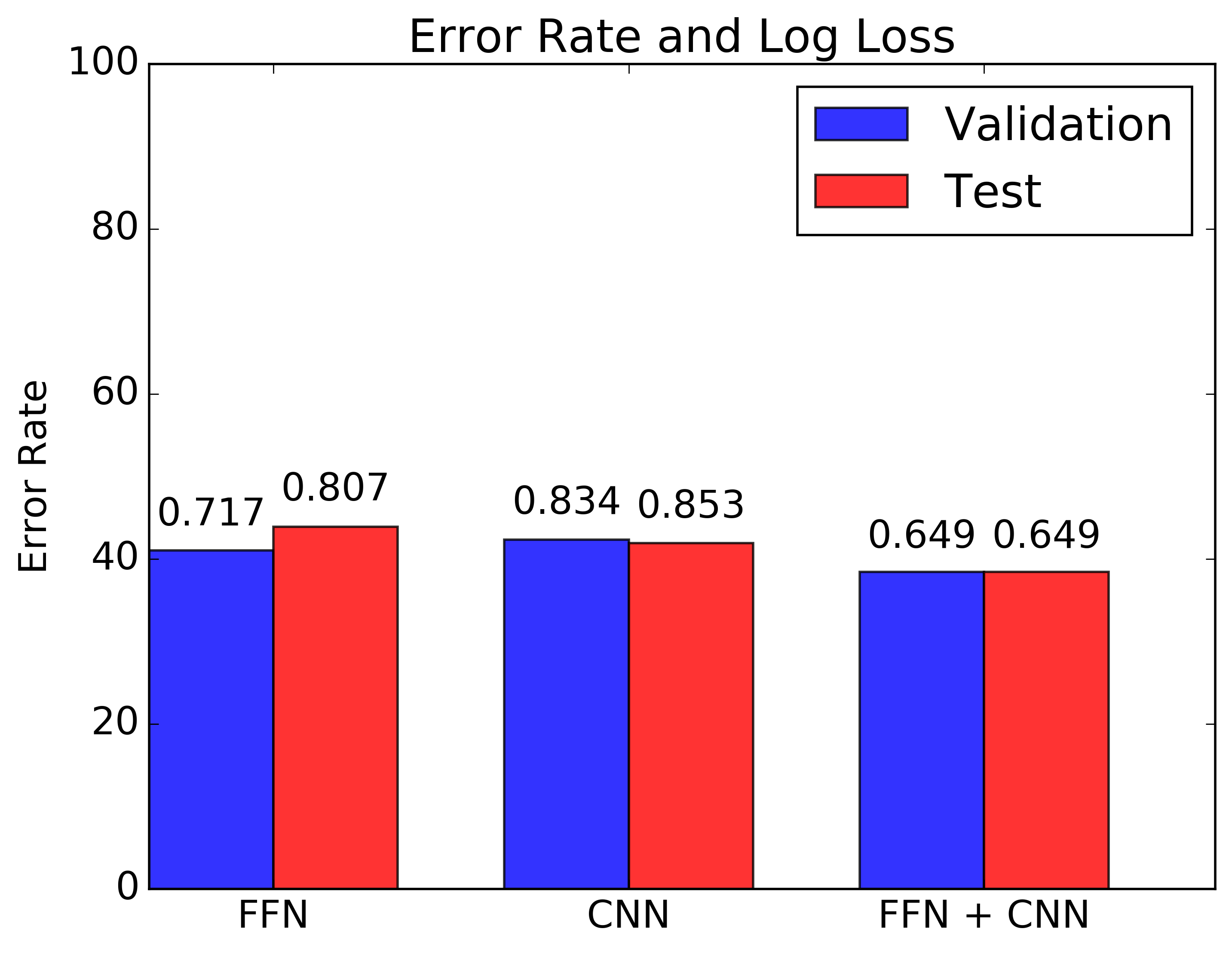}
\centering
\caption{Log loss and accuracy for FFN, CNN, and combined CNN+FFN. The combined model is the best classifier by both metrics.}
\label{log}
\end{figure}

Next, we evaluate both the accuracy and loss values for our FFN, CNN, and combined CNN+FFN models. A quick observation of the metrics represented in Figure \ref{log} shows that the final combined model is the best predictor. While the performance of the eleven channel images is an improvement over our other proposed image representations, there remains potential progress since our FFN has only slightly lower classification accuracy on the test set. 

The remaining analyses are based on the combined CNN+FFN model. In addition to assessing the accuracy of our model, we explore a basic heat map of basketball shots based upon the raw data. At the very least, we expect that our complete model should be similar to a heat map created via raw data. We make the heat map by taking a count of shots made against shots missed within a square foot of the basketball court. Since our classification model gives probabilities of making a shot (rather than a binary variable), we take the maximum probability to create a heat map equivalent to the raw data map.  From the heat map comparison, we aim to further understand how our model is making predictions.
\begin{comment}

\begin{figure*}[ht]
\begin{multicols}{2}
    \includegraphics[width=\columnwidth]{ModelHeatAllRoles.png}\par 
    \includegraphics[width=\columnwidth]{RawHeatAllRoles.png}\par
    \end{multicols}
\caption{Heat map from model data and heat map of the raw data. The hot areas depict locations in which players are much more likely to take a shot.}
\label{heatmod}
\end{figure*}
\end{comment}

\begin{figure}[ht]
\centering
\begin{minipage}[b]{0.45\textwidth}
\includegraphics[width=\textwidth]{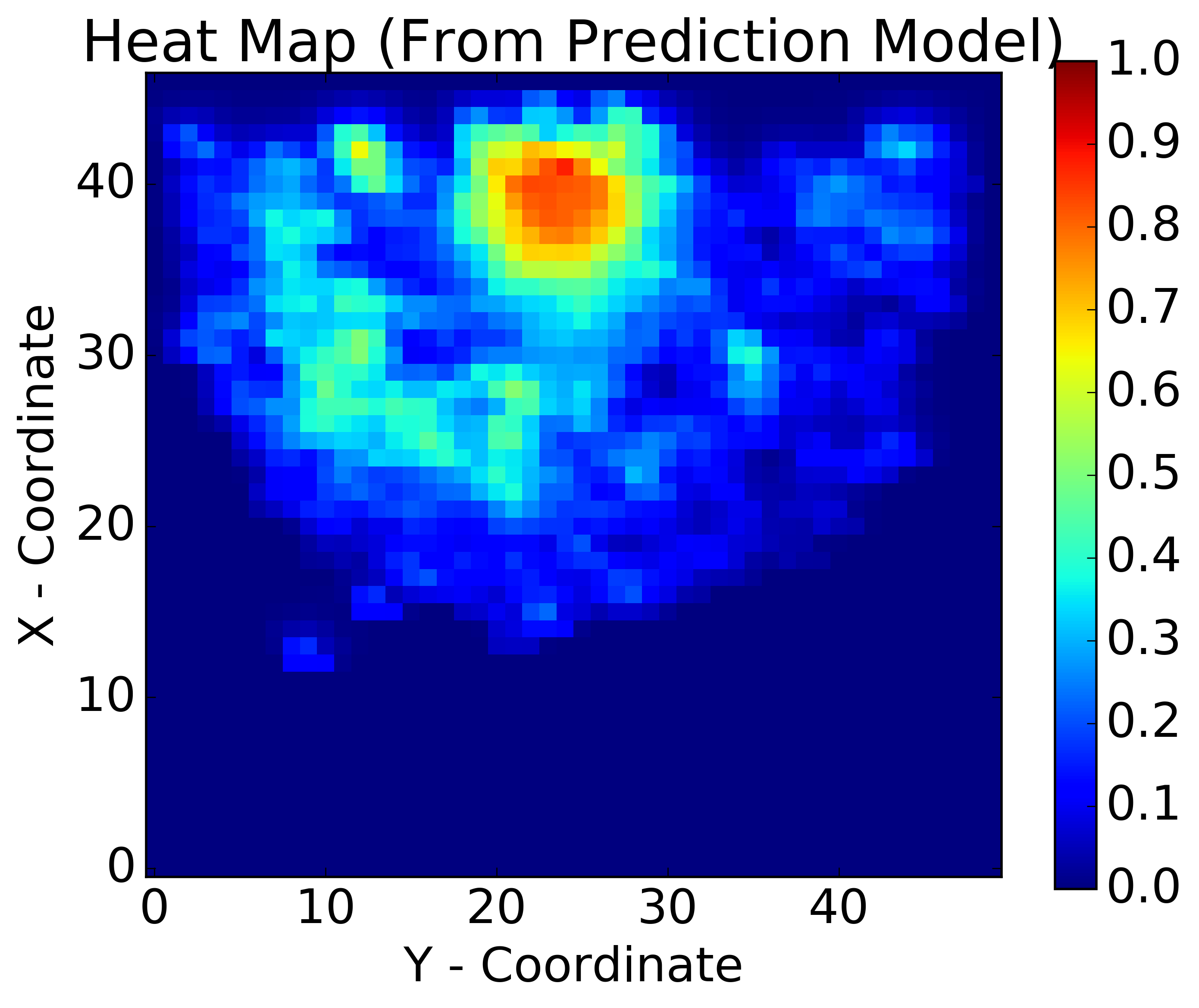}
\end{minipage}
\begin{minipage}[b]{0.45\textwidth}
\includegraphics[width=\textwidth]{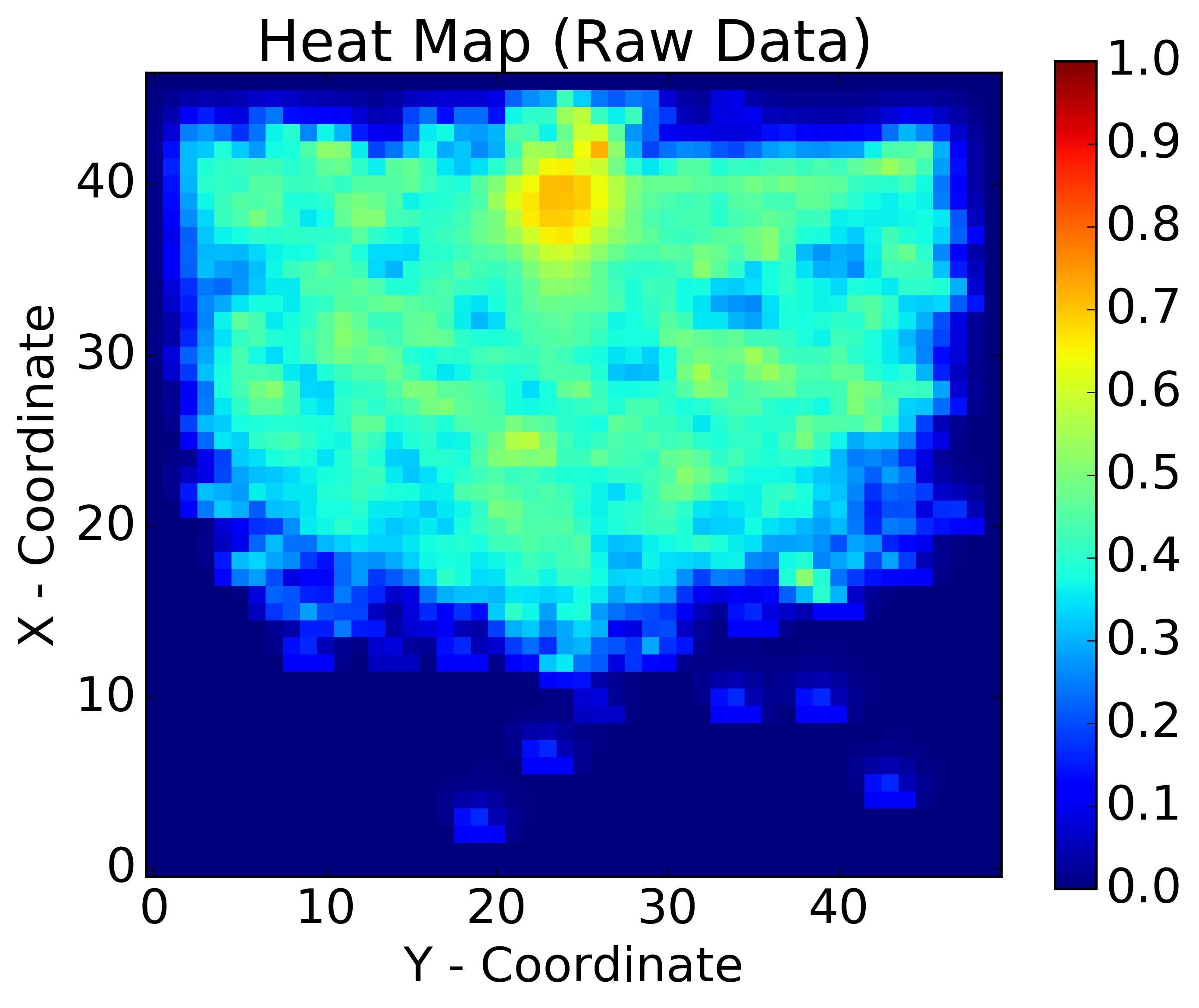}
\end{minipage}
\caption{Heat map from model data and heat map of the raw data. The hot areas depict locations in which players are much more likely to take a shot.}
\label{heatmod}
\end{figure}

In the raw data heat map, Figure \ref{heatmod}, we note that the best probability of making a shot lies on top of the basket. As we get farther back, the probability decreases with two less probable zones (lighter zones separated by a thin green strip of higher probability): one right outside the paint and another just inside the three-point line.  This means that our model is placing too much importance on the shooter being near the hoop.  The model also predicts a larger high value area surrounding the basket, which extends further into the paint of the court. However, the dead zones in the model heat map are much larger. In addition, the model over predicts near the basket while under predicting outside this area. Curiously, there are a few areas that have higher probability outside of the paint. 

\begin{figure}[ht]
\centering
\begin{minipage}[b]{0.19\textwidth}
\includegraphics[width=\textwidth]{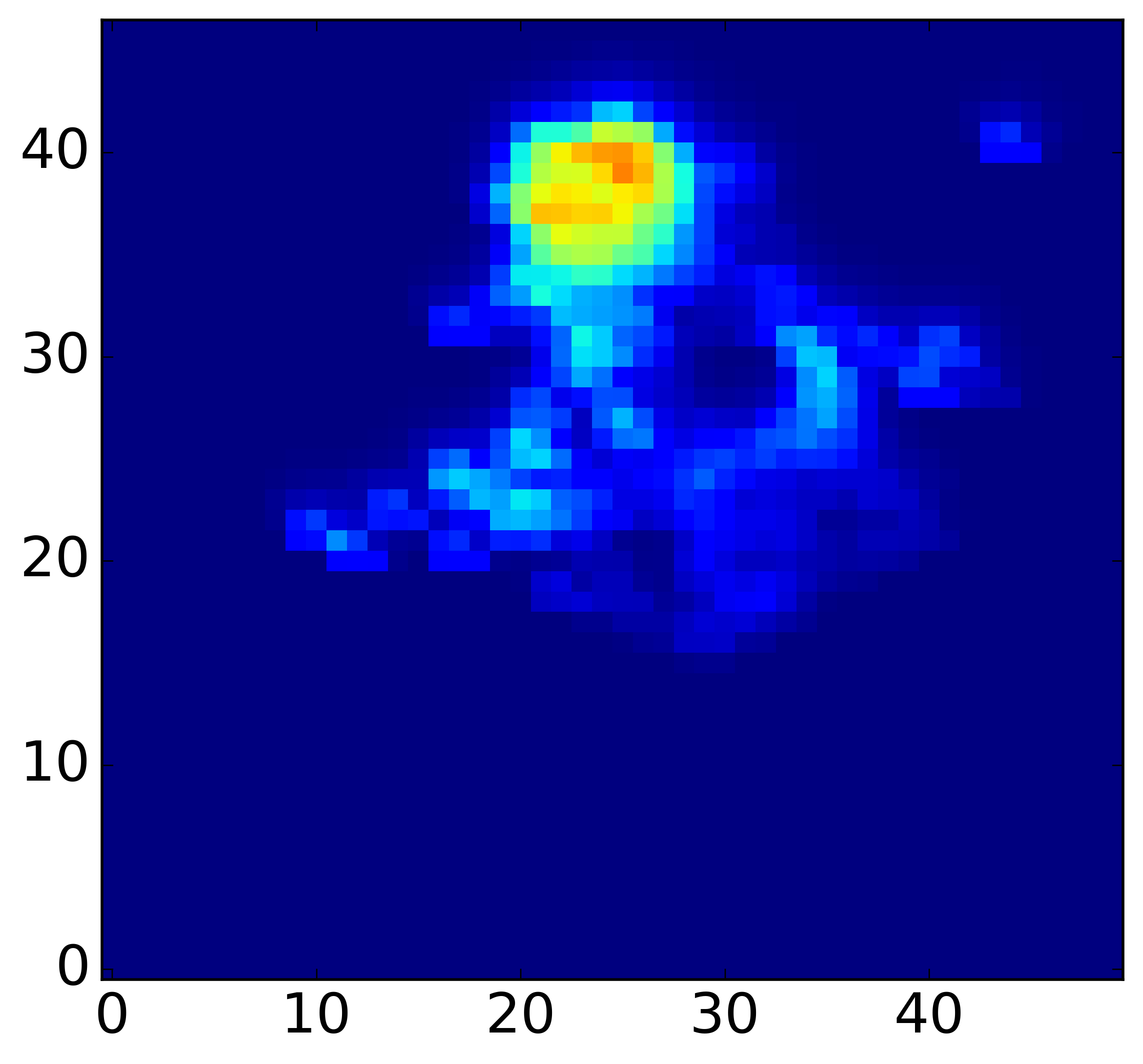}
\end{minipage}
\begin{minipage}[b]{0.19\textwidth}
\includegraphics[width=\textwidth]{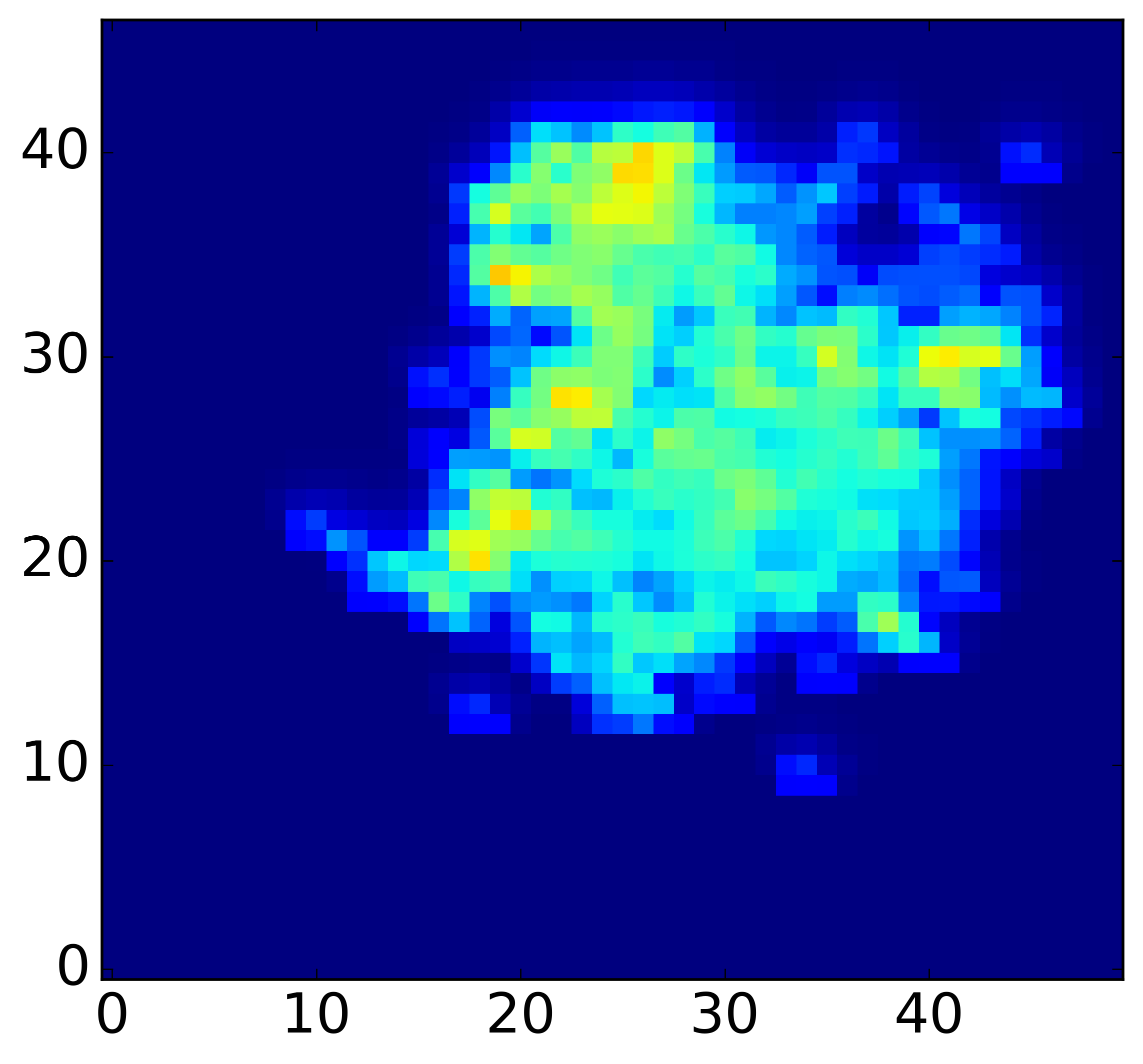}
\end{minipage}
\begin{minipage}[b]{0.19\textwidth}
\includegraphics[width=\textwidth]{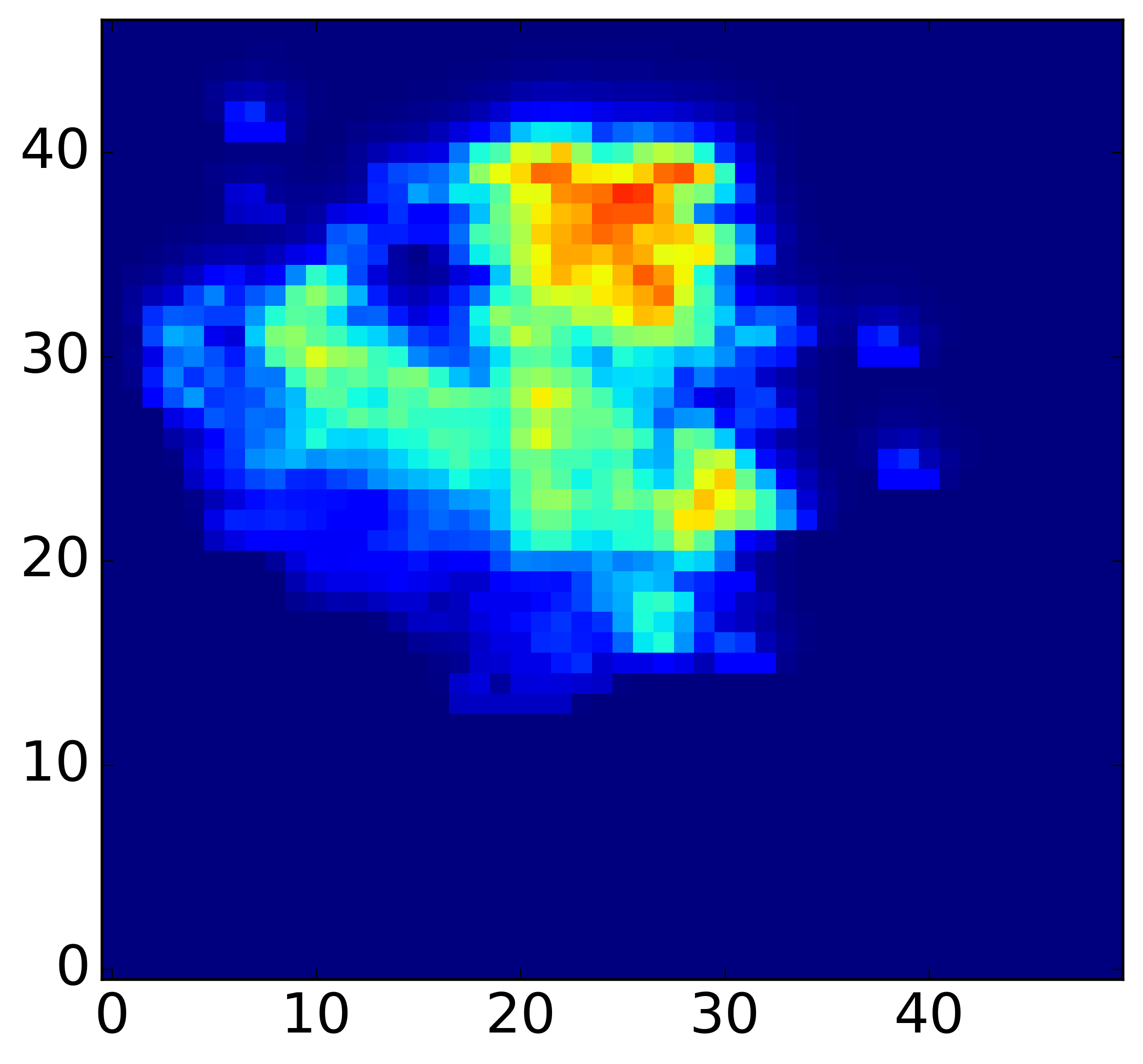}
\end{minipage}
\begin{minipage}[b]{0.19\textwidth}
\includegraphics[width=\textwidth]{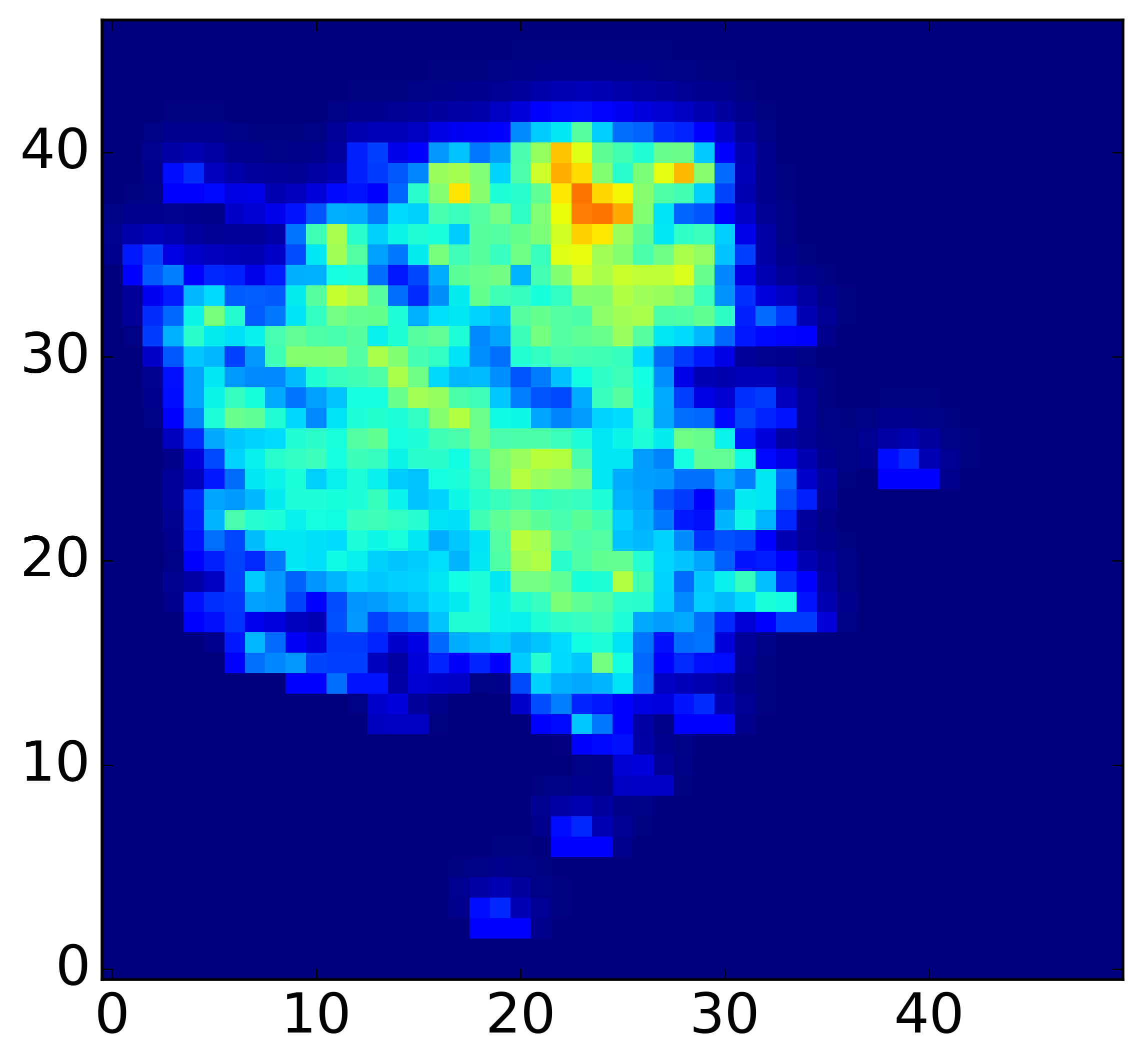}
\end{minipage}
\begin{minipage}[b]{0.19\textwidth}
\includegraphics[width=\textwidth]{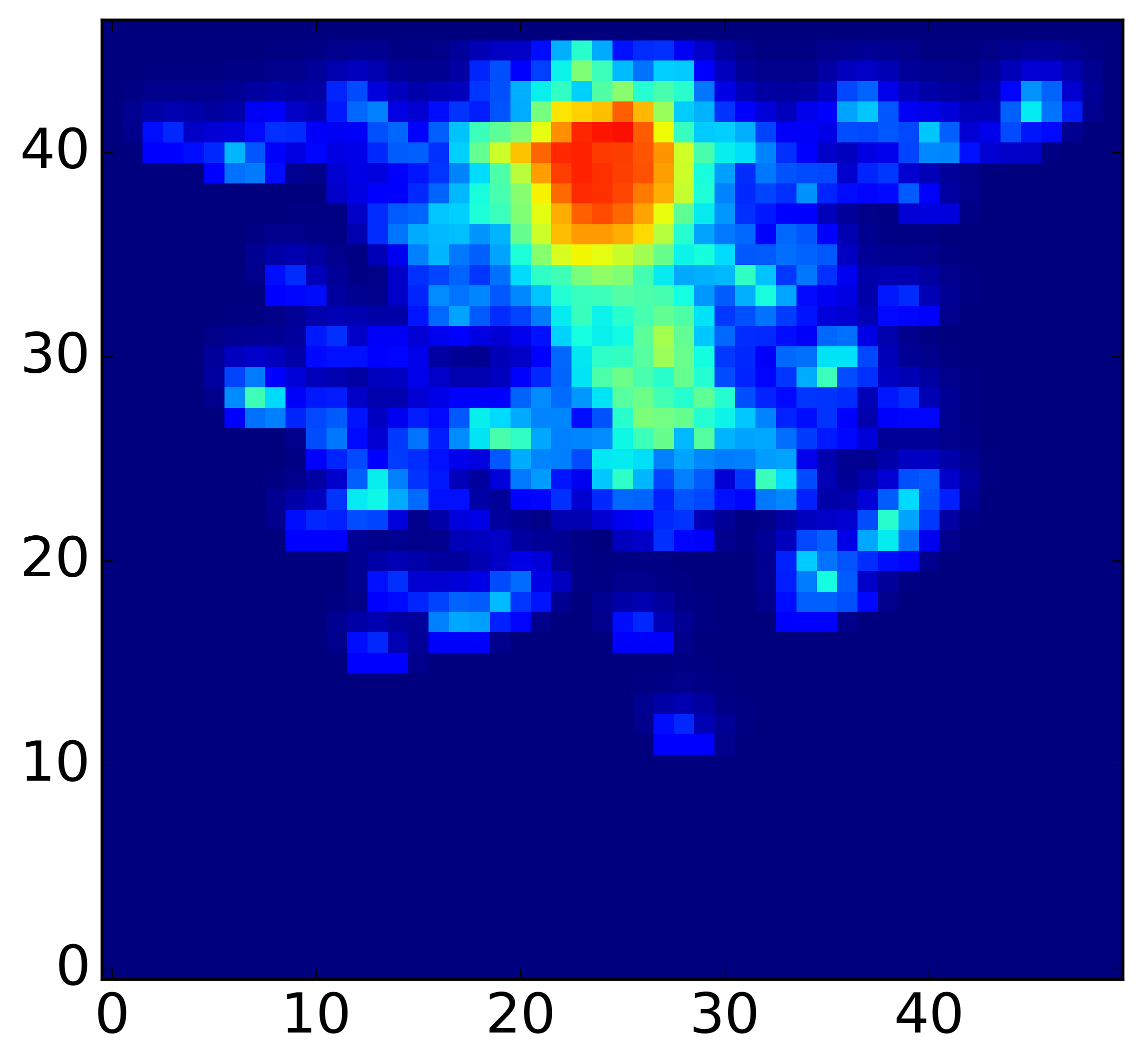}
\end{minipage}
\begin{minipage}[b]{0.19\textwidth}
\includegraphics[width=\textwidth]{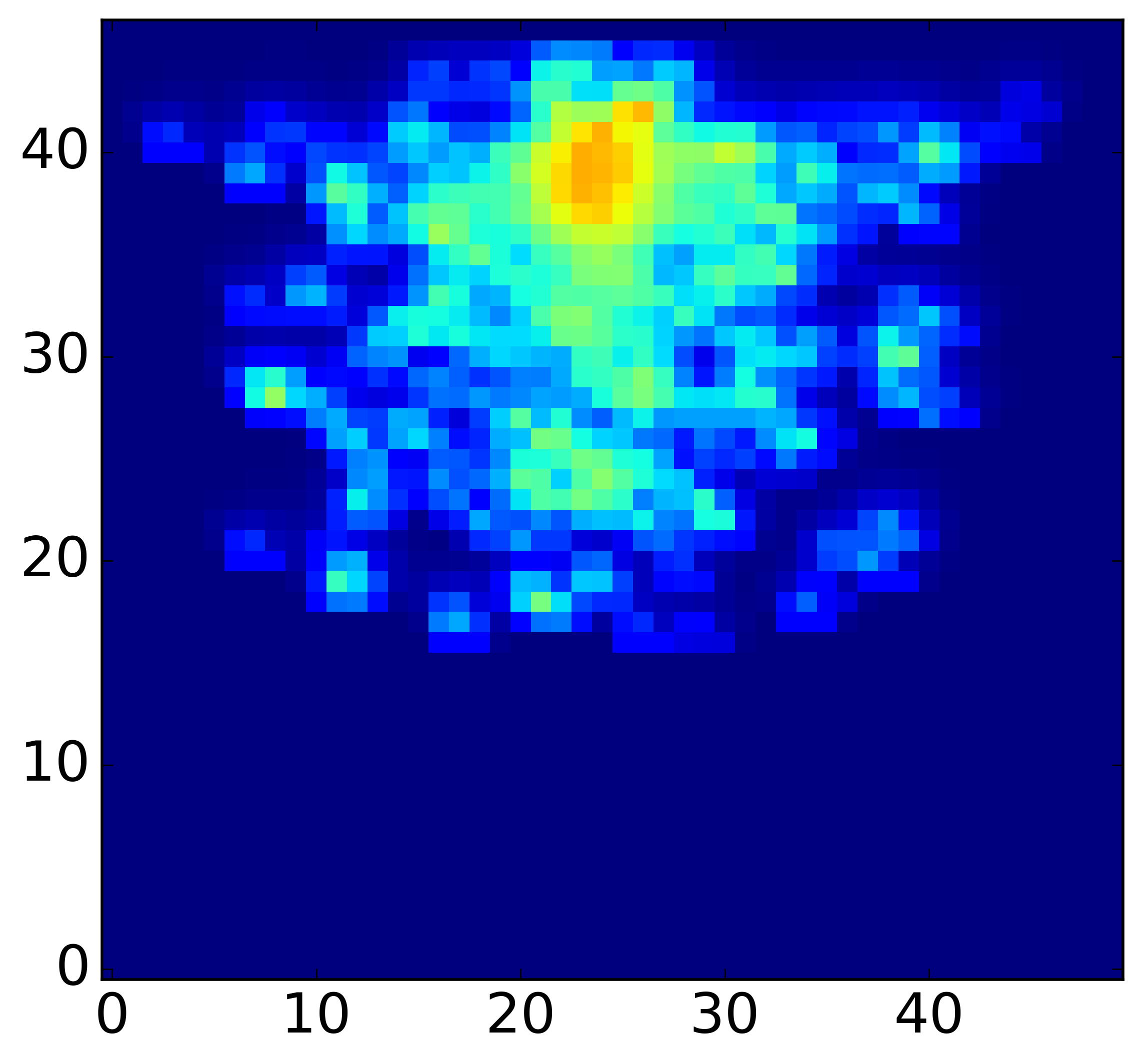}
\end{minipage}
\begin{minipage}[b]{0.19\textwidth}
\includegraphics[width=\textwidth]{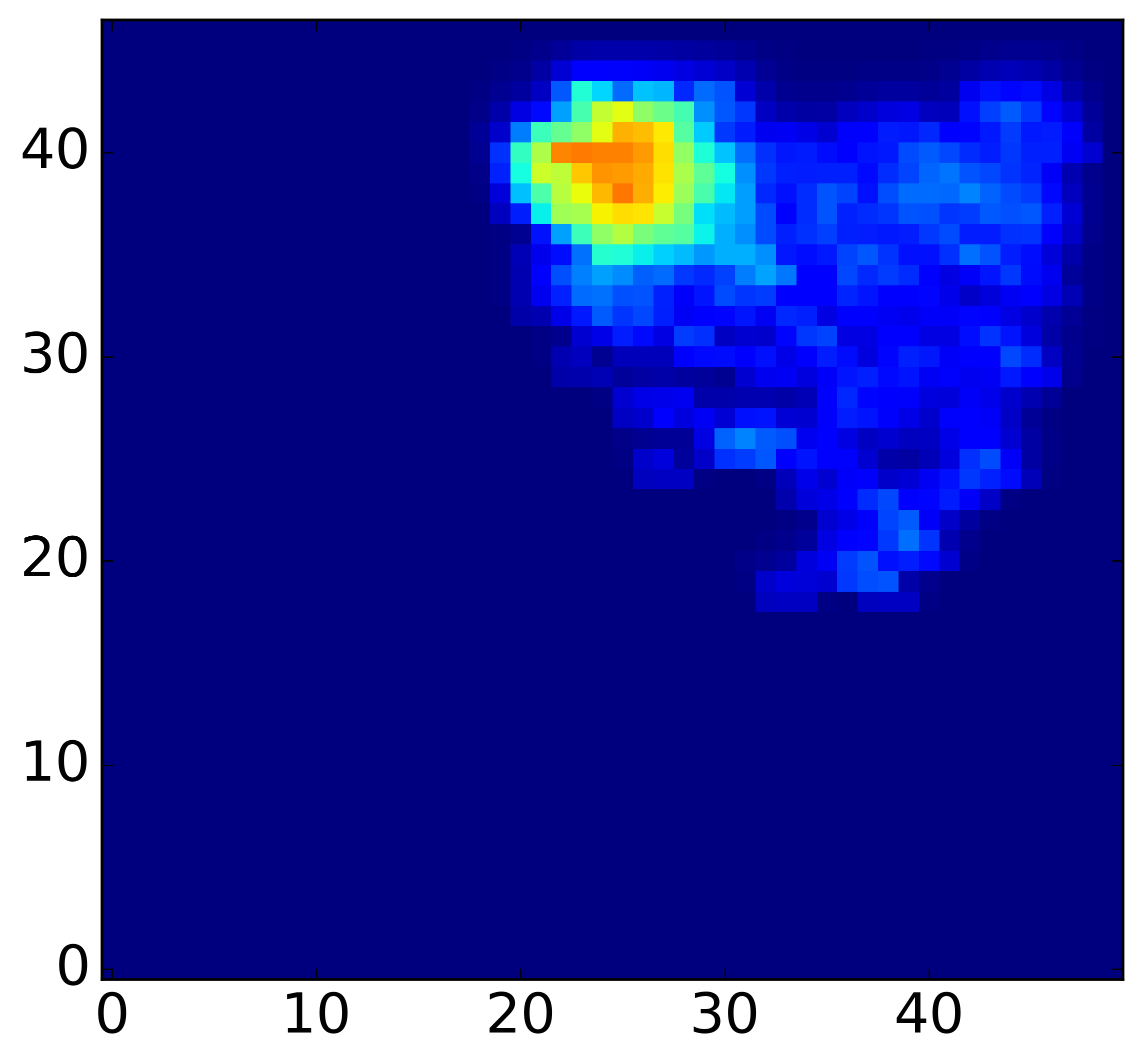}
\end{minipage}
\begin{minipage}[b]{0.19\textwidth}
\includegraphics[width=\textwidth]{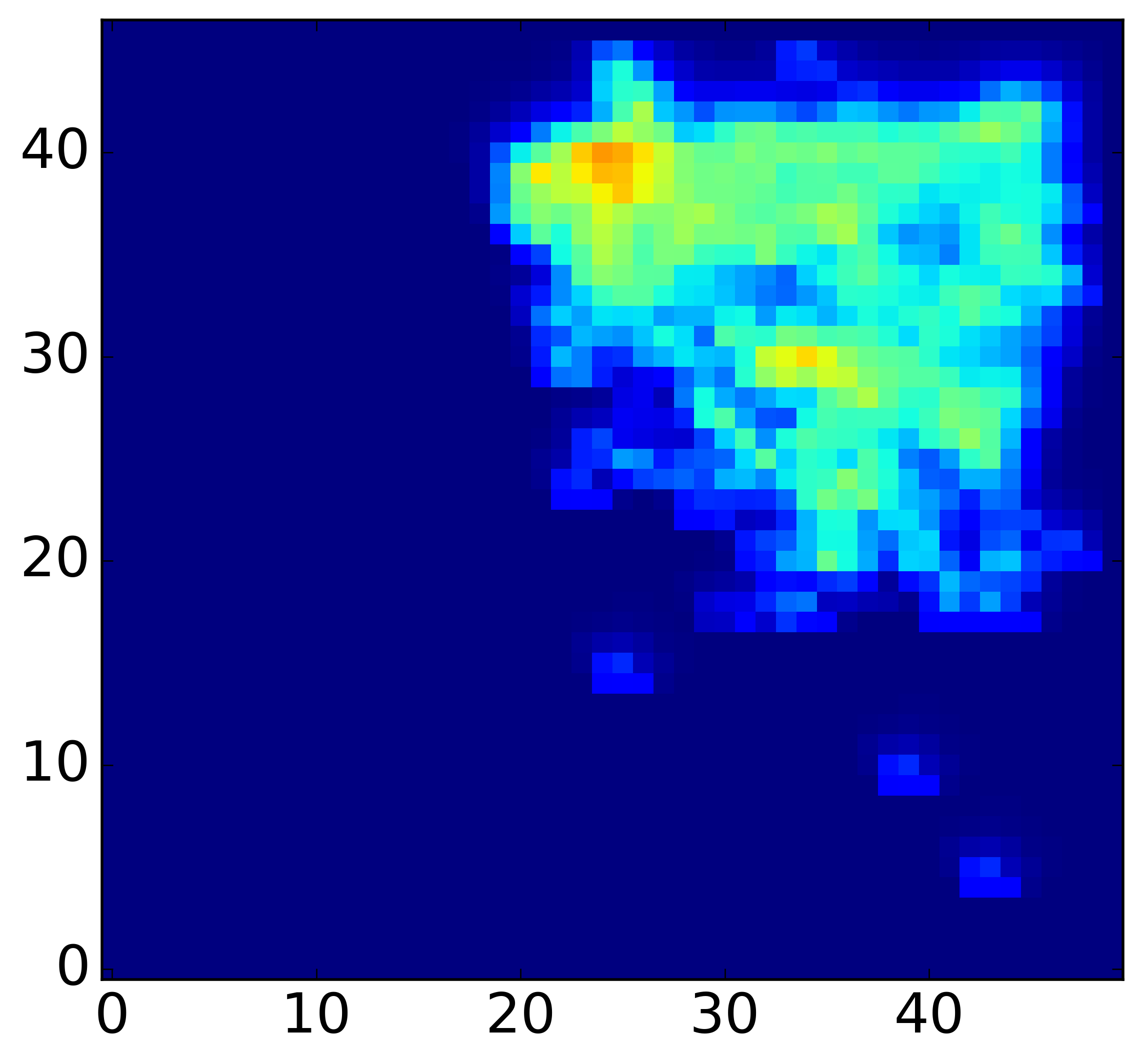}
\end{minipage}
\begin{minipage}[b]{0.19\textwidth}
\includegraphics[width=\textwidth]{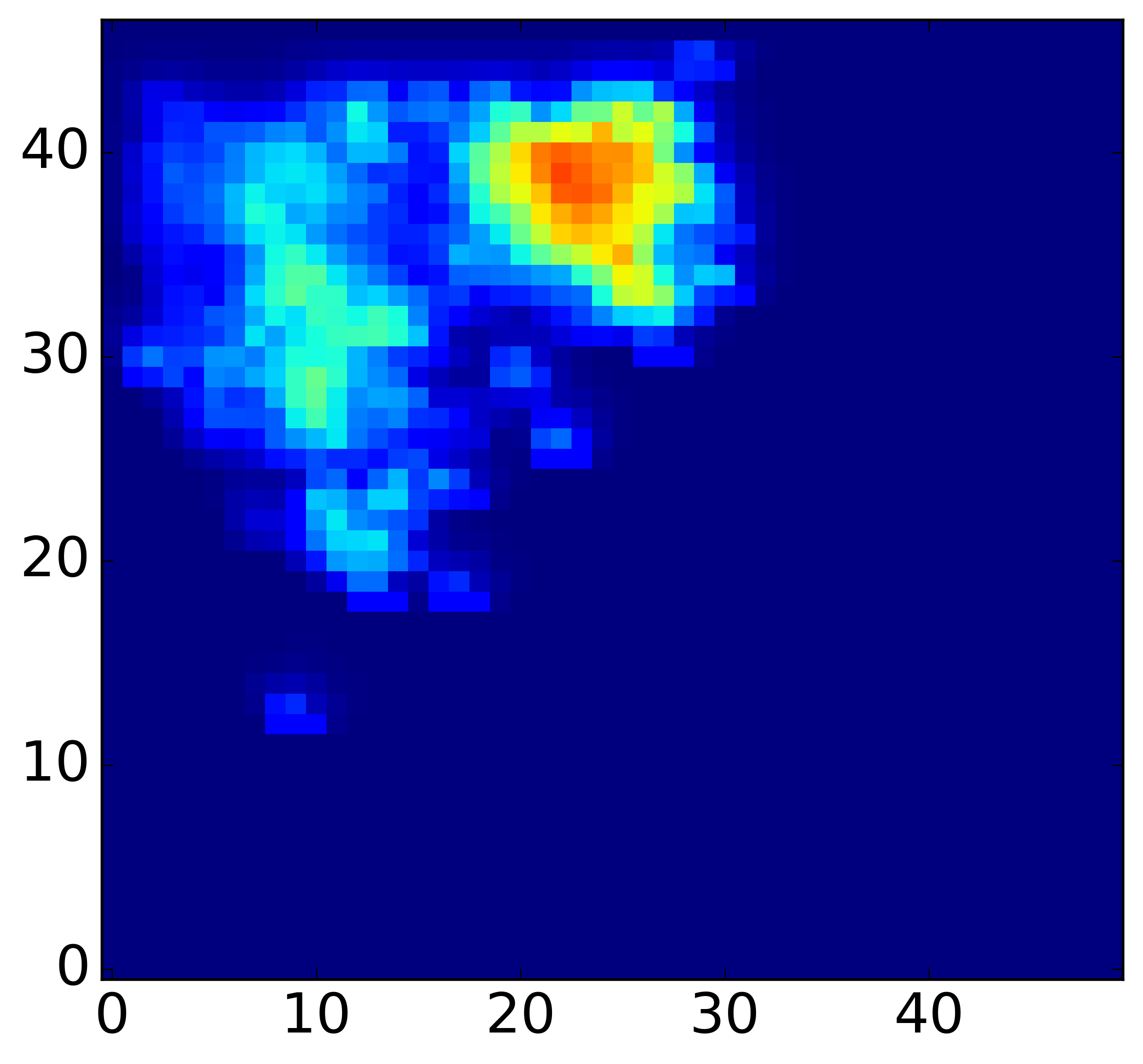}
\end{minipage}
\begin{minipage}[b]{0.19\textwidth}
\includegraphics[width=\textwidth]{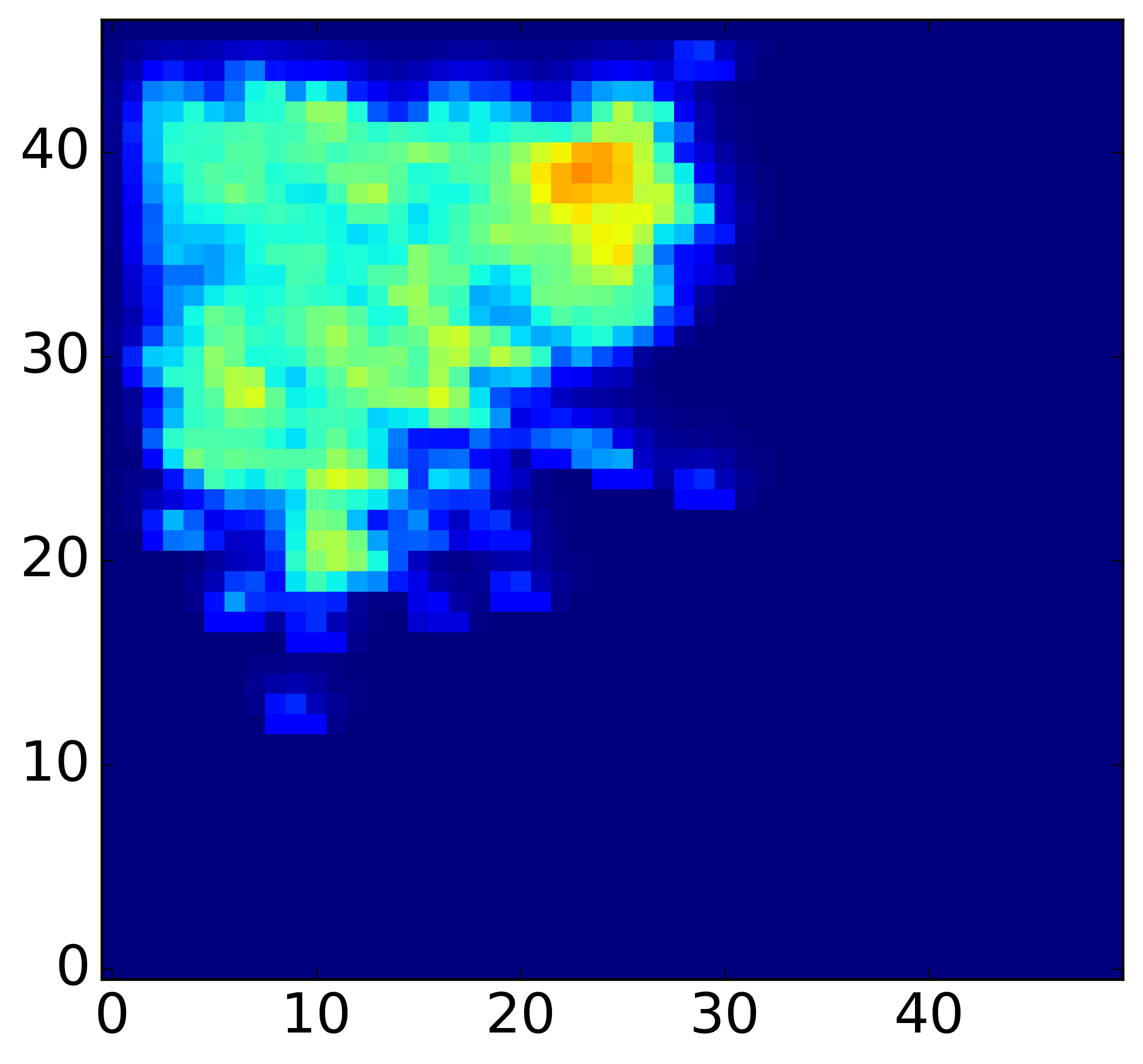}
\end{minipage}

\caption{Roles 1-5 left to right. Model results right and raw data left. The model matches best with the center (image column 3) and power forward (image column 2), but provides a larger coverage in all other positions.}
\label{heatrole}
\end{figure}

To further explore our results, we create heat maps solely based on the role of the player (to break down scoring chances by agent). As before, each role represents an offensive player. In Figure \ref{heatrole} we present a few player roles and their representative heat maps. Role 3 must be the center position from their shot selection and Role 5 is the left guard. Note that the model predicts a much smaller area of midrange scoring probability than from the raw data for Role 3. The model heat map for Role 3 strictly covers the paint, while the raw data has significantly higher shot probabilities outside of the paint. Roles 1, 4, and 5 show similar behavior in the model prediction. These maps are very heavy-handed with respect to under the basket shots with extremely small probabilities outside of the paint. The one exception is Role 2, which the model predicts has a much more likely chance of scoring outside of the paint. We observe from these heat maps that Role 2 is the reason the heat map for all in Figure \ref{heatmod} exhibits an arc of higher probability outside the hoop.  Therefore, for Role 2, our model is learning more than simply distance from the hoop for prediction in this case.  The raw data shows that shots outside the hoop are just as likely from any player, but our model argues that Role 2 is more likely to make a shot when shooting further away from the hoop.
\begin{figure*}[t]
\begin{multicols}{3}
    \includegraphics[width=\columnwidth]{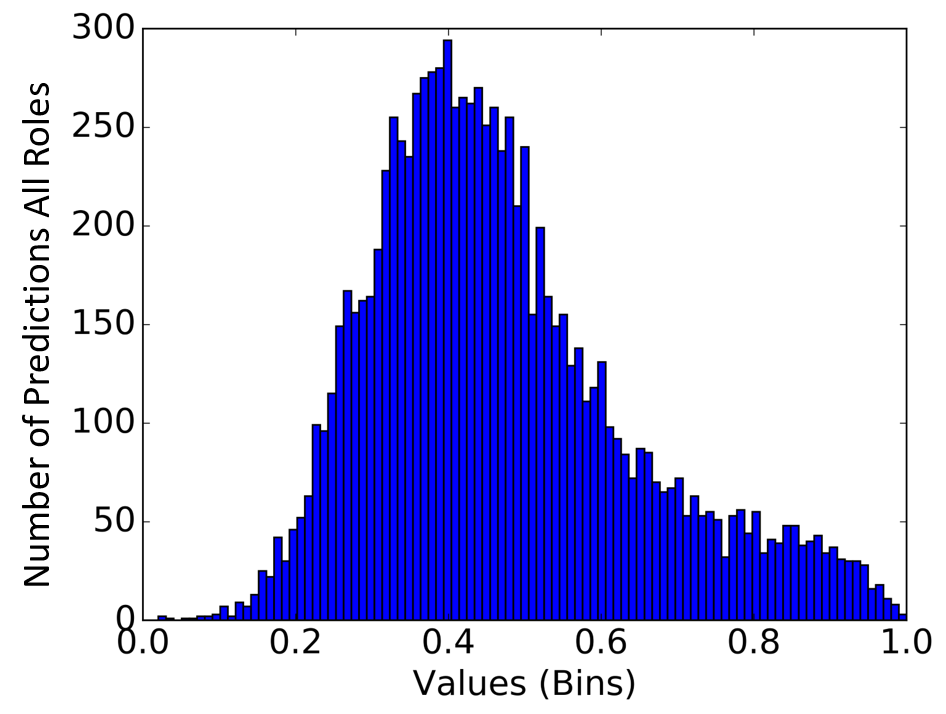}\par 
    \includegraphics[width=\columnwidth]{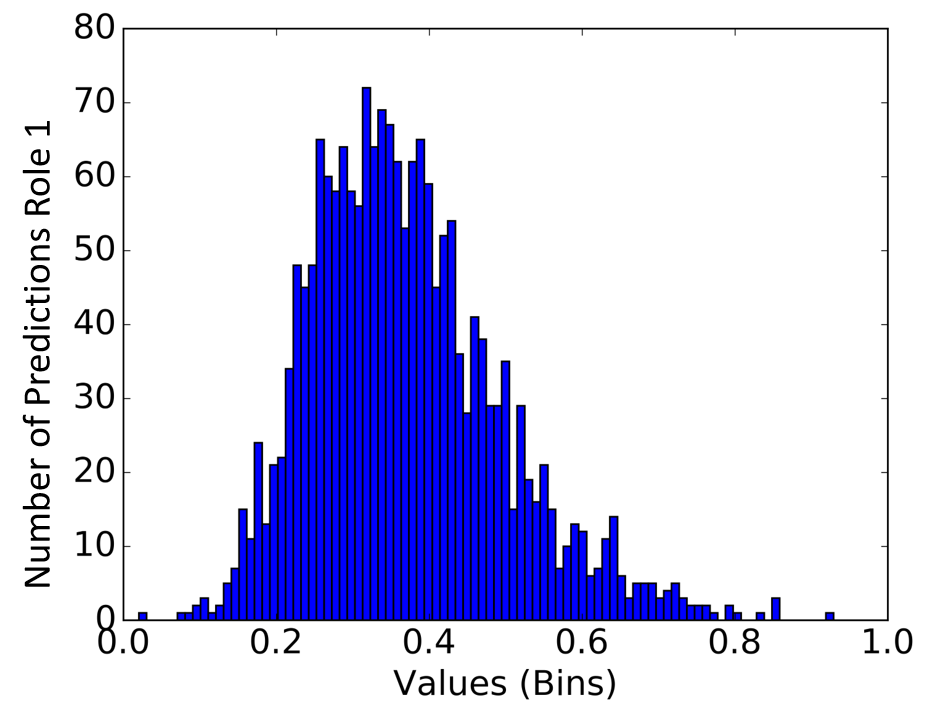}\par
        \includegraphics[width=\columnwidth]{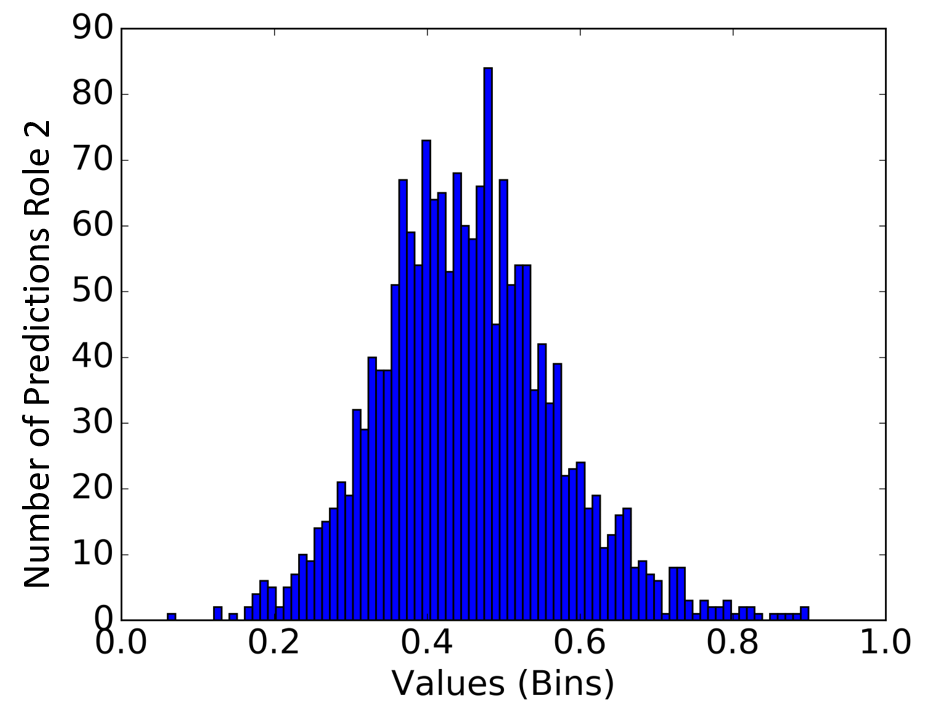}\par 
    \end{multicols}
\begin{multicols}{3}
    \includegraphics[width=\columnwidth]{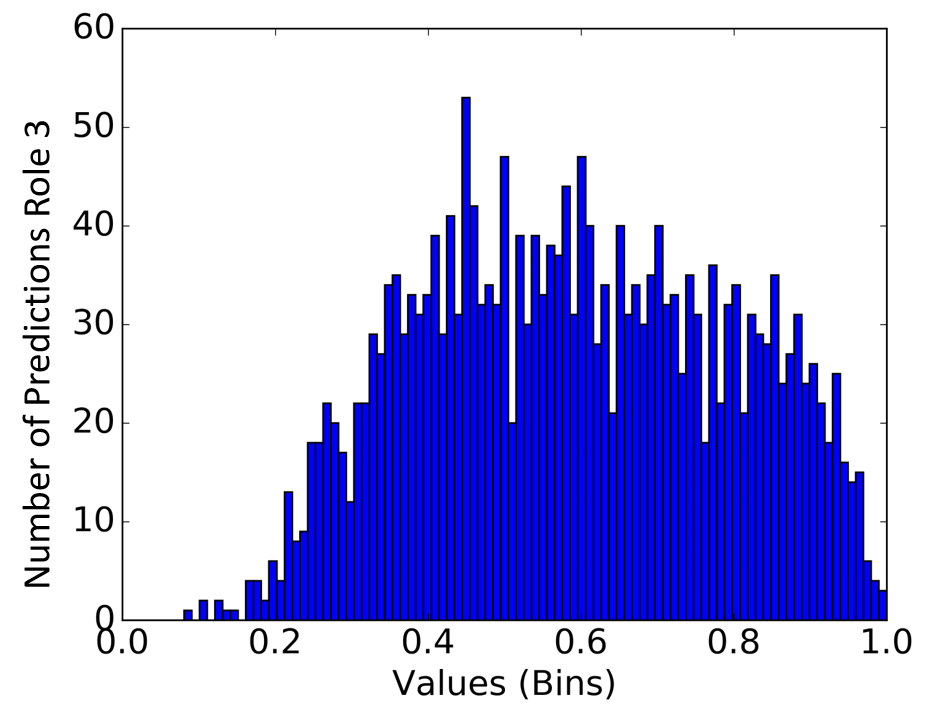}\par
        \includegraphics[width=\columnwidth]{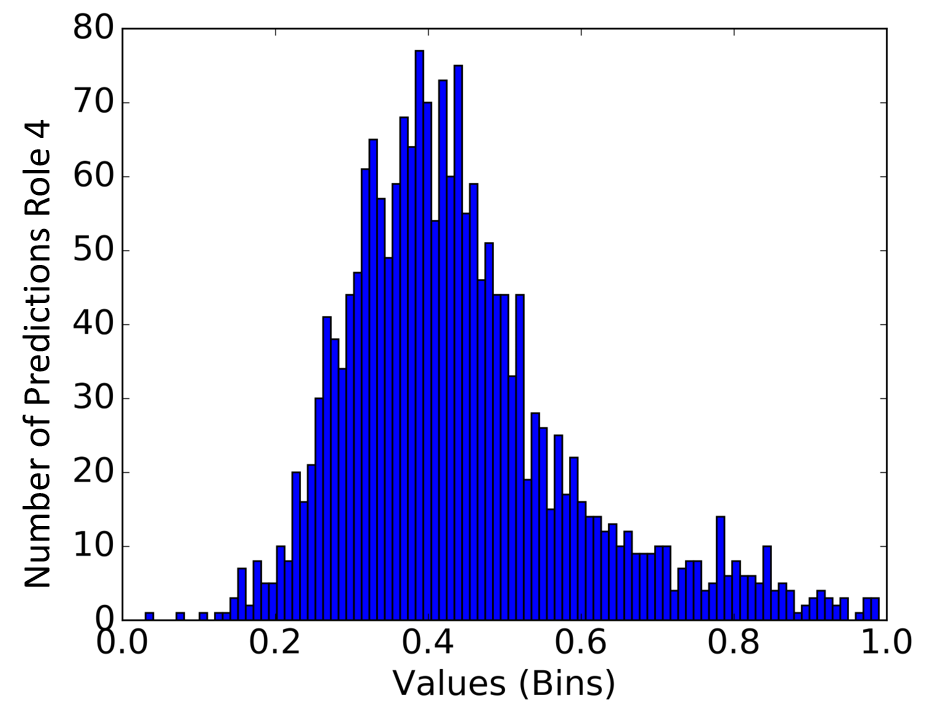}\par 
    \includegraphics[width=\columnwidth]{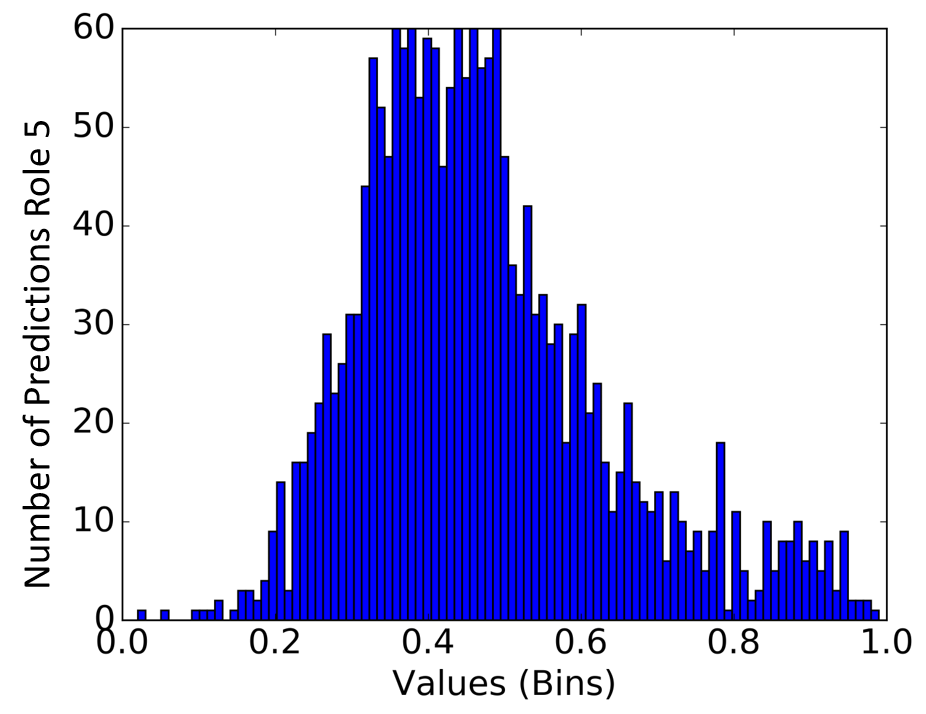}\par
\end{multicols}
\caption{From left to right and top to bottom: (1) All roles. (2) Role 1. (3) Role 2. (4) Role 3. (5) Role 4. (6) Role 5. Probabilities of shot prediction by role. Note that role 3 (the center) has the largest overall probability of making a shot.}
\label{histos}
\end{figure*}
The probabilities that the combined model predicts for each shot may also provide useful insight into the game and our model's interpretation of high versus low value shots. We create these histograms by finding which examples the model gives the highest probability as a shot made or missed by player with role $x$. We then group these examples together, and the probability of making a shot is reported in the histogram. In the histograms depicted in Figure \ref{histos} we see that most of shots have a low probability. This agrees with common basketball knowledge because many of a guard's shots are beyond the paint. On the other hand, a center remains primarily under the basket and in the paint. Therefore, many of their shots are much more likely. Watching a game live, a center getting a clean pass right under the basket often results in a made shot. In addition, most roles tend to follow the probability pattern of Role 1 except for Role 3 (the center), which has a wider distribution and higher average probability of making a shot. In addition, Role 5 tends to have a better ratio of high probability shots to low compared to Roles 1,2, and 4. A brief glance at NBA statistics agrees with this interpretation as the players with the highest shooting percentage (barring free throws) tend to be centers.

%%%%%
\begin{comment}
\begin{figure}[ht]
\centering
\begin{minipage}[b]{0.325\textwidth}
\includegraphics[width=\textwidth]{HistogramAllRoles.png}
\end{minipage} 
\begin{minipage}[b]{0.325\textwidth}
\includegraphics[width=\textwidth]{HistogramRole_1.png}
\end{minipage}
\begin{minipage}[b]{0.325\textwidth}
\includegraphics[width=\textwidth]{HistogramRole_2.png}
\end{minipage}
\begin{minipage}[b]{0.325\textwidth}
\includegraphics[width=\textwidth]{HistogramRole_3.png}
\end{minipage}
\begin{minipage}[b]{0.325\textwidth}
\includegraphics[width=\textwidth]{HistogramRole_4.png}
\end{minipage}
\begin{minipage}[b]{0.325\textwidth}
\includegraphics[width=\textwidth]{HistogramRole_5.png}
\end{minipage}
\caption{From left to right and top to bottom: (1) All roles. (2) Role 1. (3) Role 2. (4) Role 3. (5) Role 4. (6) Role 5. Probabilities of shot prediction by role. Note that role 3 (the center) has the largest overall probability of making a shot.}
\label{histos}
\end{figure}
\end{comment}
%%%%
When viewing the histograms more carefully, there are additional role dependent insights. For example, Role 1 has the lowest probabilities for shots made compared to all other roles. Although the general shape is the same as for Role 2, the predictions are shifted significantly to the left. We also note that overall, most of the shots fall into the 45\% category of shot being made, which is aligned with general basketball knowledge. Since these histograms are predictions of the average basketball player in each role, the takeaway message is that unless you have an ace shooter (Lebron James or Steph Curry), a team should focus on ball movement to give the center an open shot rather than relying on outside shooters.

\begin{figure*}[ht]
\begin{multicols}{3}
    \includegraphics[width=\columnwidth]{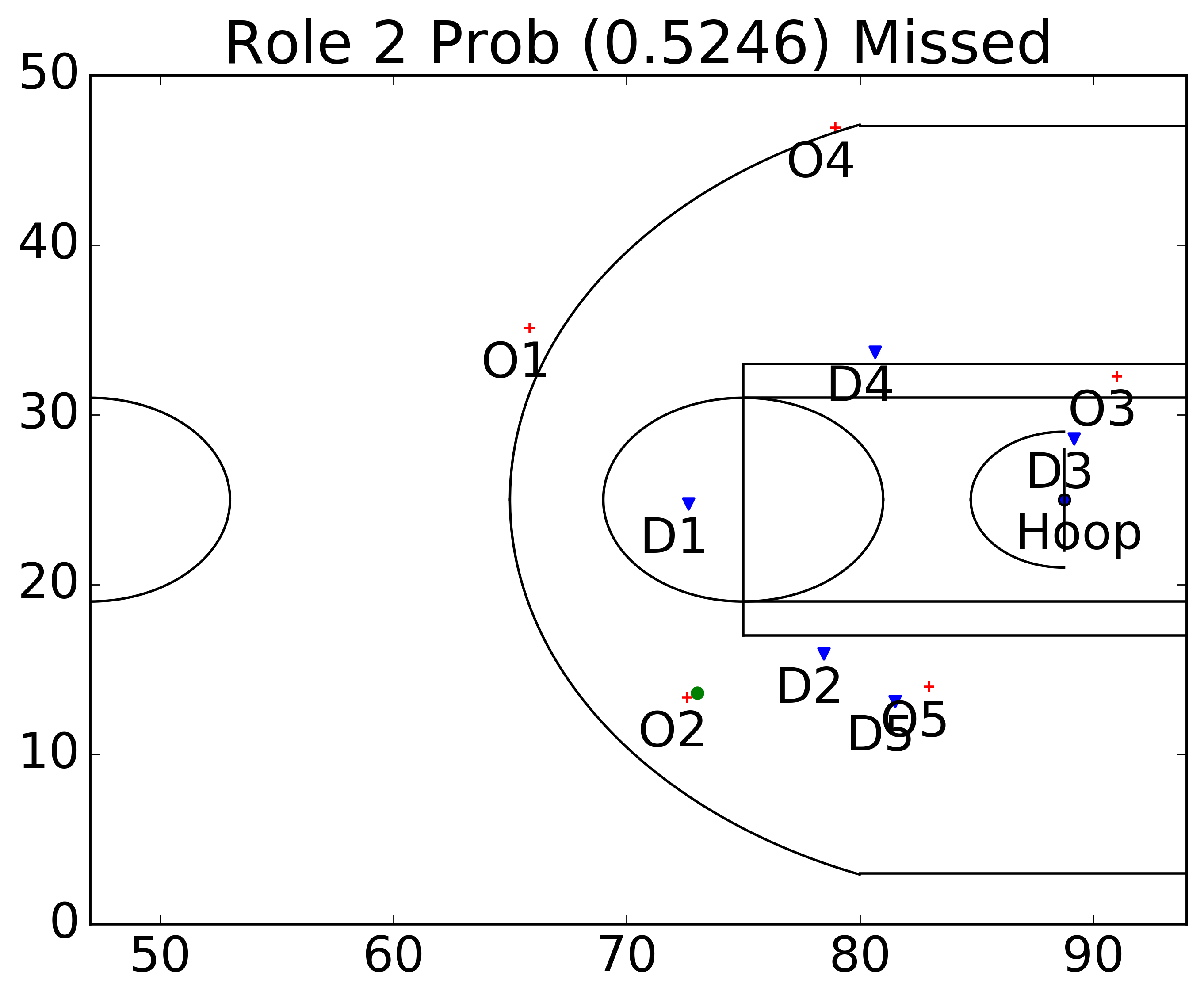}\par 
    \includegraphics[width=\columnwidth]{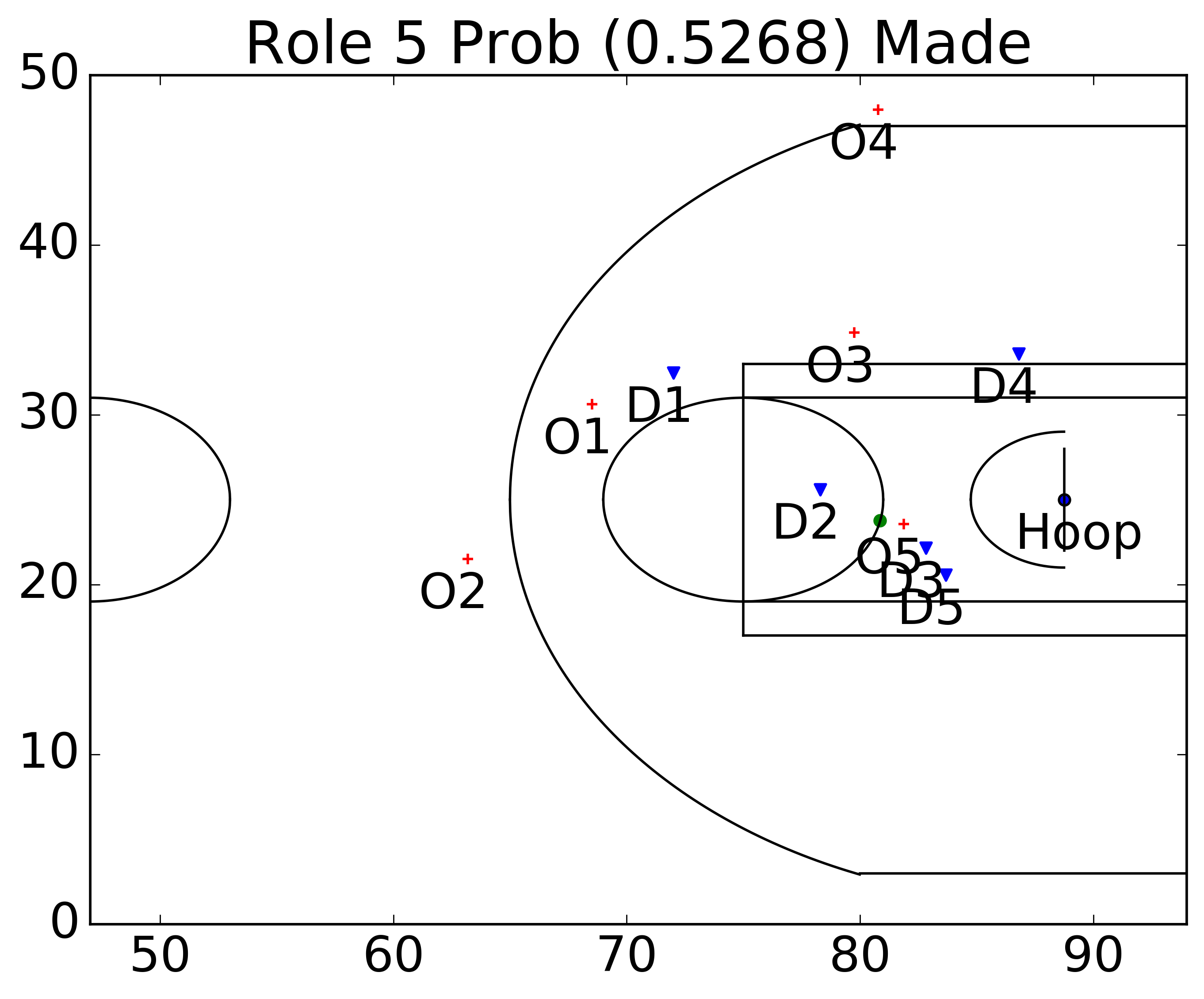}\par
        \includegraphics[width=\columnwidth]{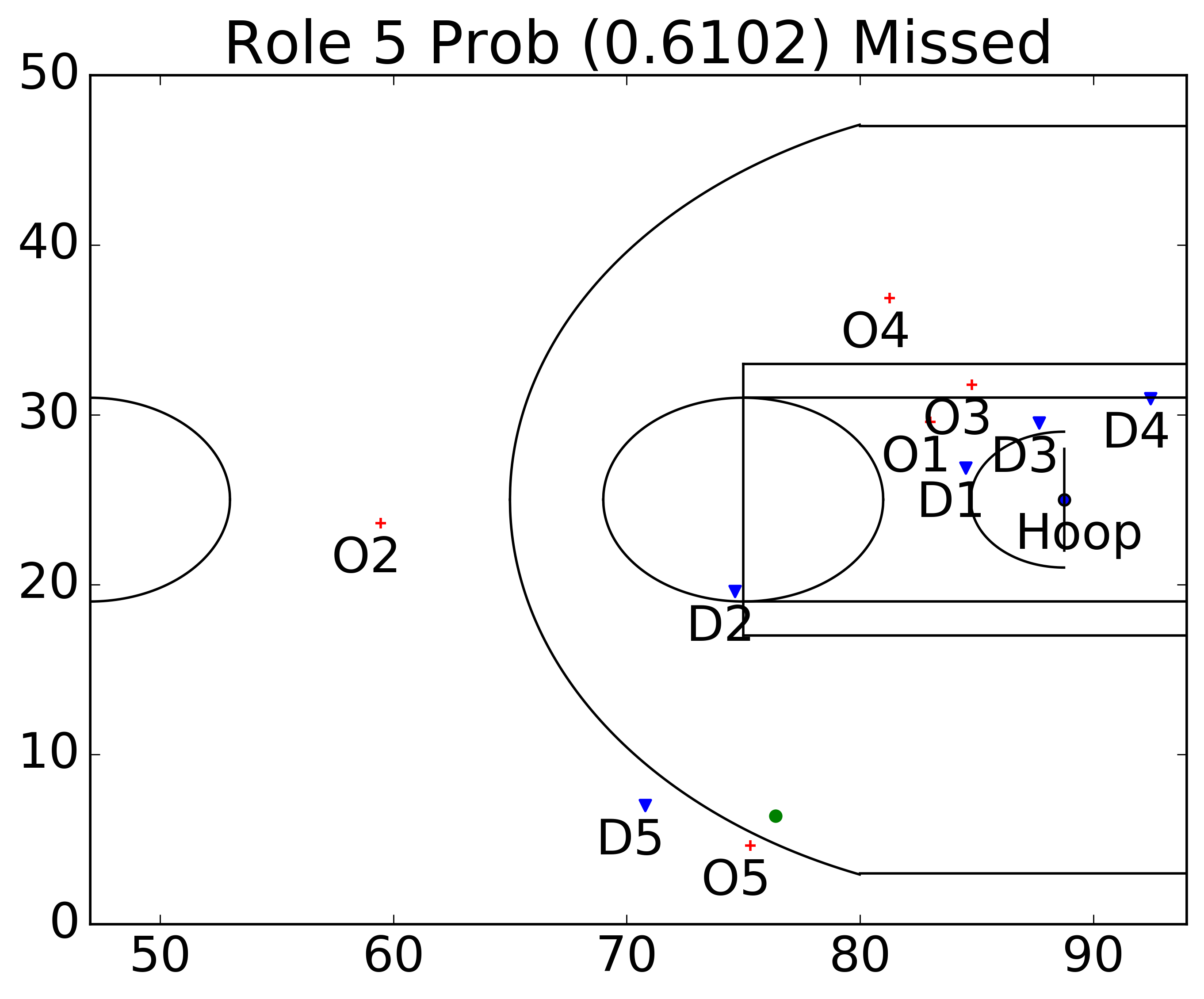}\par 
    \end{multicols}
\begin{multicols}{3}
    \includegraphics[width=\columnwidth]{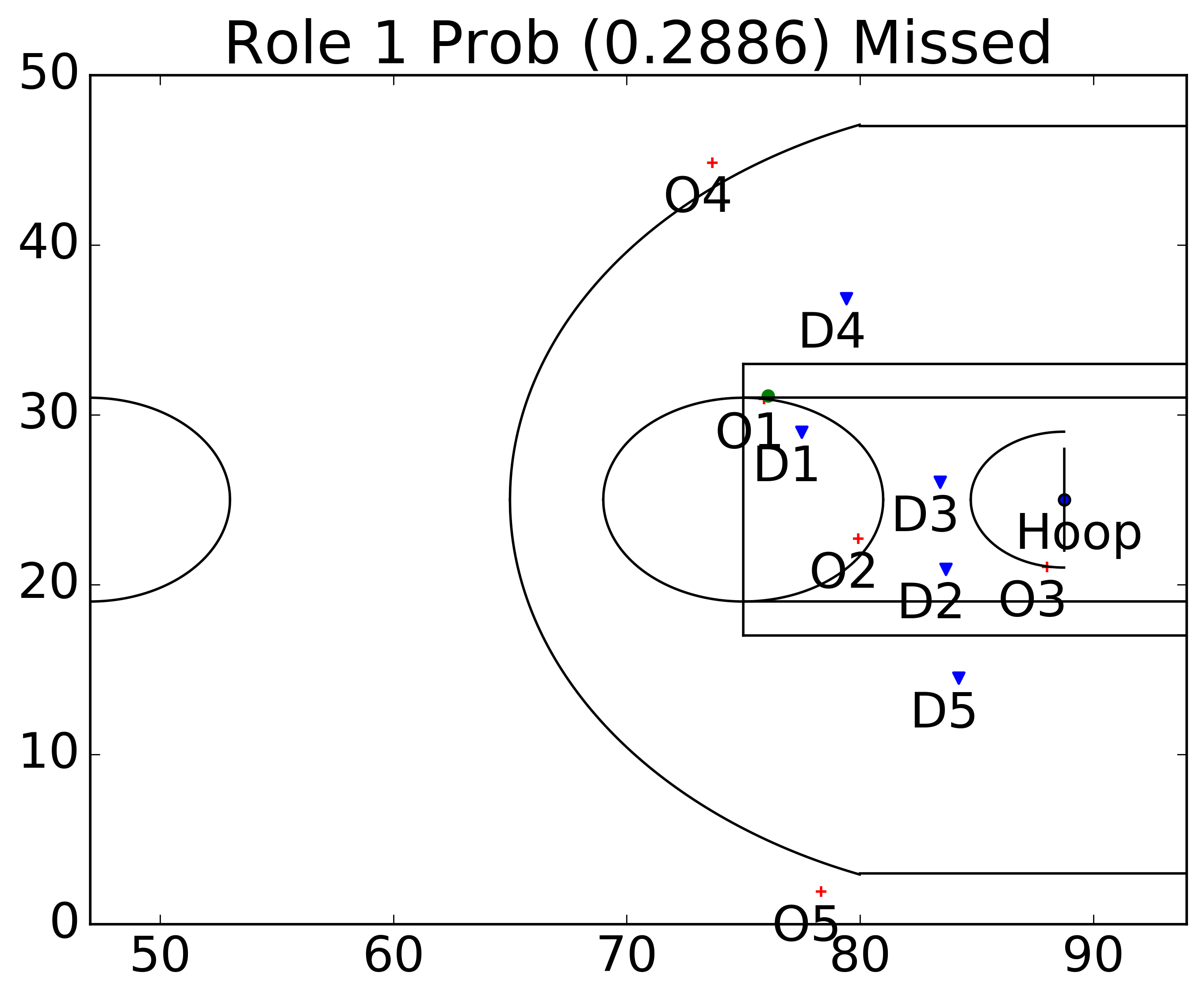}\par
        \includegraphics[width=\columnwidth]{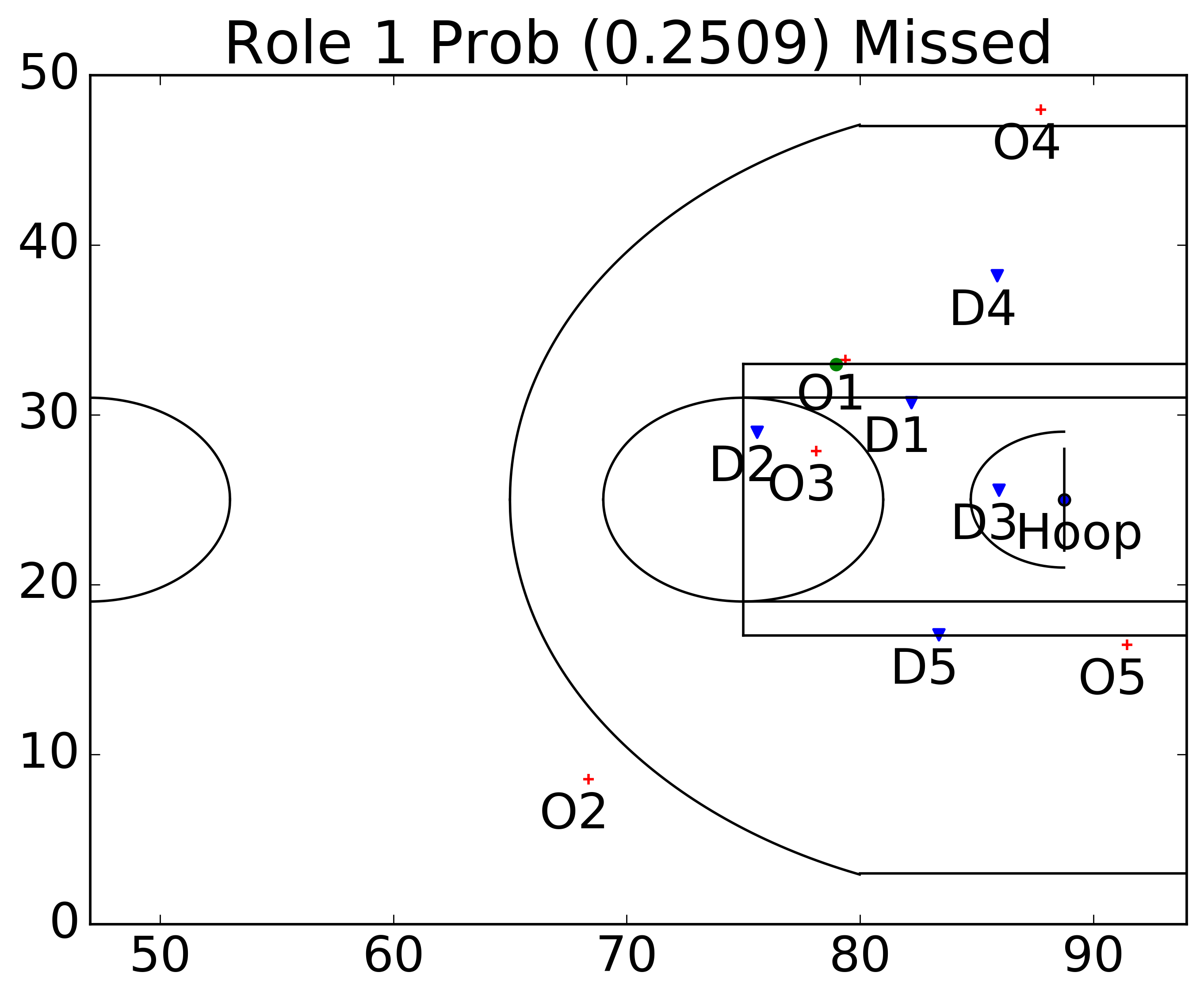}\par 
    \includegraphics[width=\columnwidth]{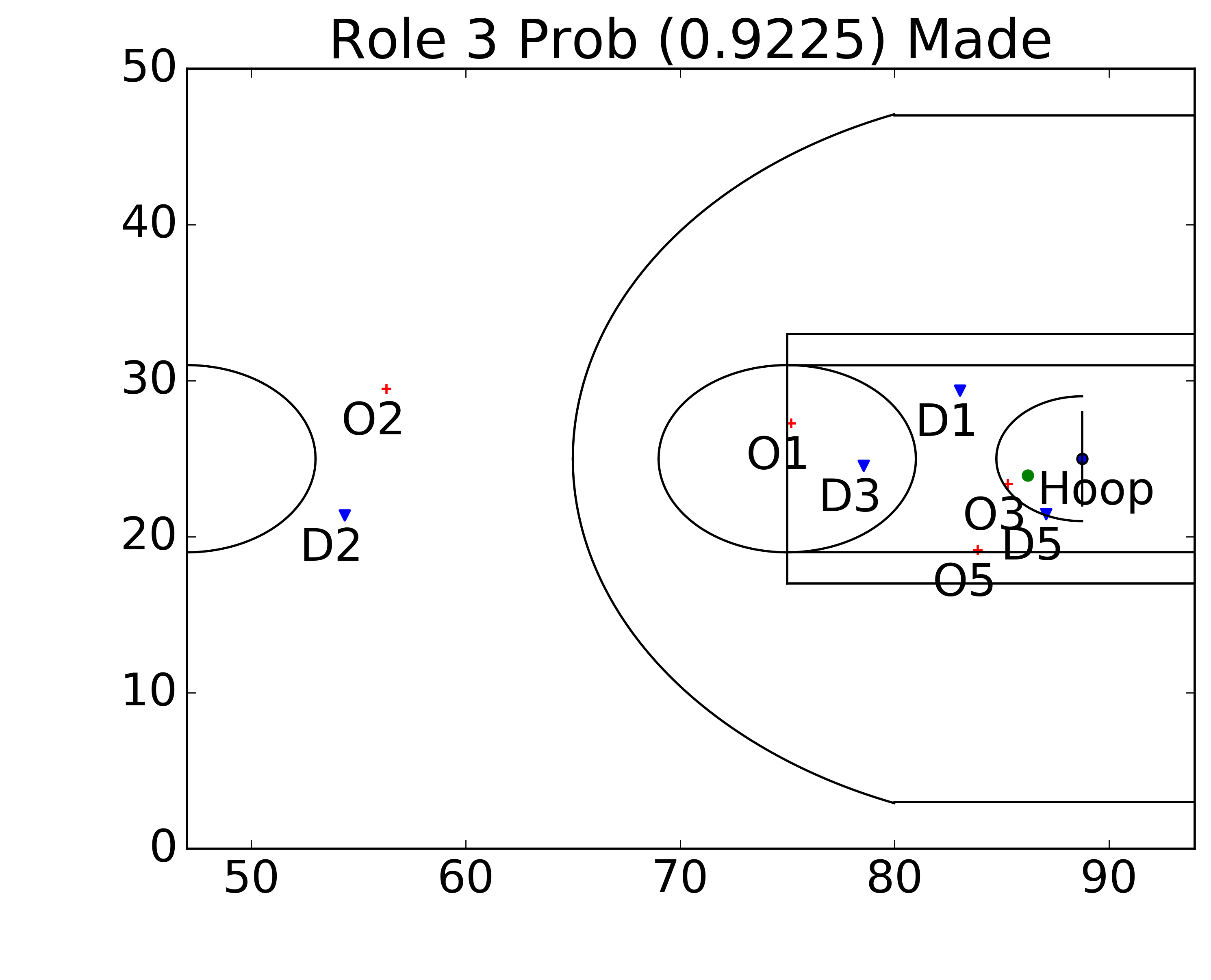}\par
\end{multicols}
\caption{Locations of offensive (O) and defensive (D) players on the court at the time of the shot displaying the model's prediction probabilities. Most of the predictions are reasonable when considering shot location and defensive positioning.}
\label{bballexamps}
\end{figure*}

We also exhibit Figure \ref{bballexamps} to provide additional visual context for our model probabilities. These figures depict the final positions of all players on the court at the time of the shot. We can assess how ``open" the shot maker is at the time of the shot and the relative position of both the offensive and defensive players. The offensive players are blue, defensive players red, and the ball is green. Each offensive and defensive player has the letter ``O" and ``D" respectively followed by a number signifying the role of that player at that time. There are times when the model makes some questionable predictions. For example, the three-point shot that is exhibited in the top right of Figure \ref{bballexamps} with a 0.610 probability is much too high. Unguarded, we do not expect three-point shots to be made more than 50\% of the time. Thankfully, these examples are very rare in our model. For the most part, three-point shots are rated extremely low by the model garnering probabilities of less than 20\%.
\begin{comment}
\begin{figure}[h]
\centering
\begin{minipage}[b]{0.32\textwidth}
\includegraphics[width=\textwidth,frame]{ShotProbability_0_Role2.png}
\end{minipage}
\begin{minipage}[b]{0.32\textwidth}
\includegraphics[width=\textwidth,frame]{ShotProbability_2_Role5.png}
\end{minipage}
\begin{minipage}[b]{0.32\textwidth}
\includegraphics[width=\textwidth,frame]{ShotProbability_1_Role5.png}
\end{minipage}
\begin{minipage}[b]{0.32\textwidth}
\includegraphics[width=\textwidth,frame]{ShotProbability_0_Role1.png}
\end{minipage}
\begin{minipage}[b]{0.32\textwidth}
\includegraphics[width=\textwidth,frame]{ShotProbability_7_Role1.png}
\end{minipage}
\begin{minipage}[b]{0.32\textwidth}
\includegraphics[width=\textwidth,frame]{ShotProbability_11_Role3.png}
\end{minipage}
\caption{Locations of offensive (O) and defensive (D) players on the court at the time of the shot displaying the model's prediction probabilities. Most of the predictions are reasonable when considering shot location and defensive positioning.}
\label{bballexamps}
\end{figure}
\end{comment}

In addition to three-point shots having a generally low probability, shots that are well-covered by defenders have a much lower probability of success. This is an unsurprising well-known result, but it does add validity to our model. In addition, shots that are open and close to the basket are heavily favored in our model. For example, in the bottom right picture, Role 3 has a very good chance of making a wide-open shot with the defense well out of position.

As noted in Figure \ref{log}, the FFN classifies with nearly the same accuracy as the CNN; however, the combined CNN+FFN model is the best classifier. Therefore, the CNN must learn features that elude the FFN.

We next explore our CNN model by creating images that result in maximum activation in the CNN (\cite{erhan2009visualizing}). The goal of creating these images is to find the features in our images that the CNN model learns. The process is similar to a reverse of training a neural network.  First, we take an already trained CNN and a randomly created image. Then without changing the weights $\theta^*$ of the CNN that feed into filter $i$ at layer $\ell$, we use gradient ascent with respect to the image $x$ that yield maximum value to activation $a_{i}^{\ell}$. To make the image from the filter we solve: $x^* = \arg\max_{x} a_{i}^{\ell}(\theta^{*},x)$.

\begin{figure*}[ht]
\begin{multicols}{4}
    \includegraphics[width=\columnwidth]{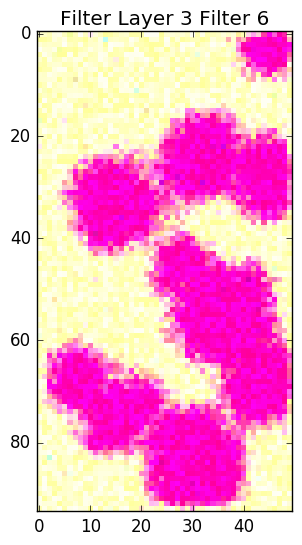}\par 
    \includegraphics[width=\columnwidth]{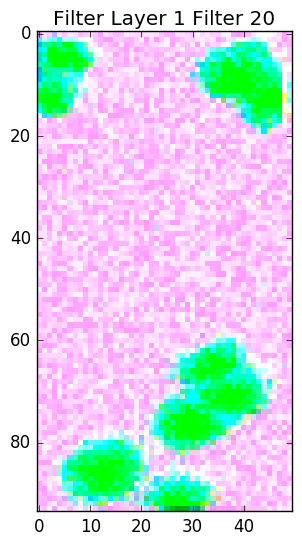}\par
    \includegraphics[width=\columnwidth]{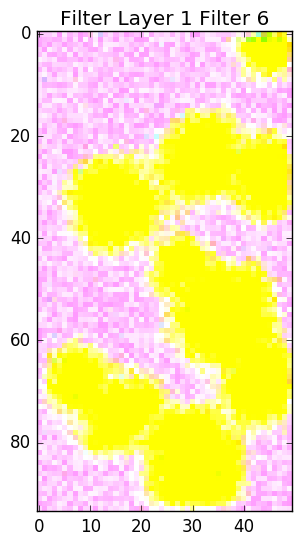}\par
    \includegraphics[width=\columnwidth]{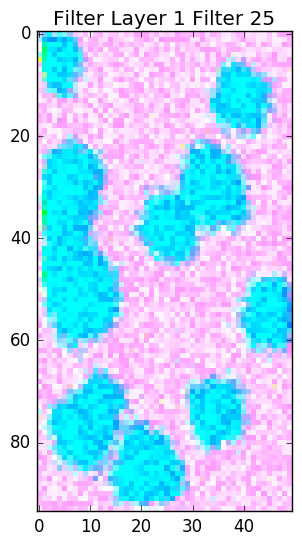}\par
    \end{multicols}

\caption{From left to right: (1) Filter 6 in convolution layer 3 of the offensive players (2) Filter 20 in convolution layer 1 of the ball (3) Filter 6 in convolution layer 1 of the offense and ball (4) Filter 25 in convolution layer 1 of the defense. Images that yield maximum activation using a trained CNN.}
\label{filters1}
\end{figure*}

\begin{comment}

\begin{figure}[ht]
\centering
\begin{minipage}[b]{0.12\textwidth}
\includegraphics[width=\textwidth]{Filter1_Rev.png}
\end{minipage}
\begin{minipage}[b]{0.12\textwidth}
\includegraphics[width=\textwidth]{Filter2_Rev.png}
\end{minipage}
\begin{minipage}[b]{0.12\textwidth}
\includegraphics[width=\textwidth]{Filter3_Rev.png}
\end{minipage}
\begin{minipage}[b]{0.12\textwidth}
\includegraphics[width=\textwidth]{Filter4_Rev.png}
\end{minipage}
\caption{From left to right: (1) Filter 6 in convolution layer 3 of the offensive players (2) Filter 20 in convolution layer 1 of the ball (3) Filter 6 in convolution layer 1 of the offense and ball (4) Filter 25 in convolution layer 1 of the defense. Images that yield maximum activation using a trained CNN.}
\label{filters1}
\end{figure}

\end{comment}

In Figure \ref{filters1}, we present four images from four distinct filters in our CNN model (several other filters result in white noise).  The images in the figure are an RGB representation of the eleven channel images we create with the maximum activation method. Since we know which agent is represented by each channel, we let green, red, and blue represent the ball, offense, and defense, respectively. We implement this transformation into the RGB space for qualitative and quantitative assessments when compared to historical data of the ball, offense, and defense. 

The first and third images of Figure \ref{filters1} are nearly identical; however, while the first image displays primarily offensive areas, the third image presents the same areas but with ball (green) information as well. The second image shows that the filter is attempting to identify ball (green) activity, and the fourth image is a filter identifying defensive (blue) activity. We did not expect filters to look for offensive, defensive, or ball activity near the top of the court (away from the hoop) since we ensure that the offense always shoots towards the bottom of the image.
\begin{figure*}[ht]
\begin{multicols}{3}
    \includegraphics[width=\columnwidth]{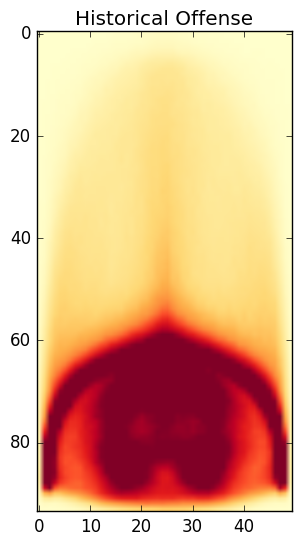}\par 
    \includegraphics[width=\columnwidth]{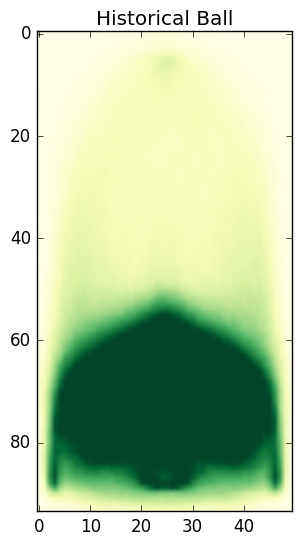}\par
        \includegraphics[width=\columnwidth]{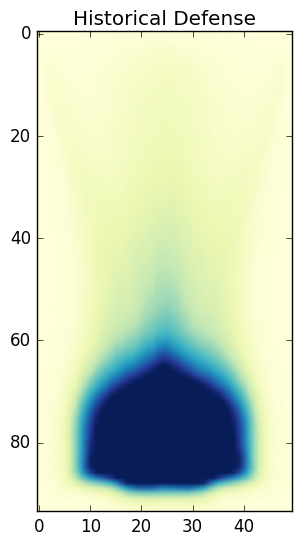}\par 
    \end{multicols}

\caption{From left to right: (1) Historical offense data (2) Historical ball data (3) Historical defense data. Figures on the court of hot spots for the offense, ball, and defense at the time of a shot. Note that the defense is much more tightly packed around the hoop than the offense.}
\label{histmap}
\end{figure*}

\begin{comment}

\end{c}
\begin{figure*}
  \centering
  \begin{tabular}{c c c}
  \includegraphics[width=1.2in]{JournalOff.png} &
  \includegraphics[width=1.2in]{JournalBall.png} &
  \includegraphics[width=1.2in]{JournalDef.png} \\
  a.~Historical offense data A & b.~Historical ball data & c.~Historical defense data
  \end{tabular}
  \caption{Figures on the court of hot spots for the offense, ball, and defense at the time of a shot. Note that the defense is much more tightly packed around the hoop than the offense.}
  \label{histmap}
\end{figure*}
\end{comment}
\begin{comment}
\begin{figure}[h]
\centering
\begin{minipage}[b]{0.24\textwidth}
\includegraphics[width=\textwidth]{JournalOff.png}
\end{minipage}
\begin{minipage}[b]{0.24\textwidth}
\includegraphics[width=\textwidth]{JournalBall.png}
\end{minipage}
\begin{minipage}[b]{0.24\textwidth}
\includegraphics[width=\textwidth]{JournalDef.png}
\end{minipage}
\caption{From left to right: (1) Historical offense data (2) Historical ball data (3) Historical defense data. Figures on the court of hot spots for the offense, ball, and defense at the time of a shot. Note that the defense is much more tightly packed around the hoop than the offense.}
\label{histmap}
\end{figure}
\end{comment}
Figure \ref{histmap} is a historical representation of the locations of all offensive (including the ball) and defensive agent locations at the time of the shot for our five second plays. For ease of comparison, we retain our qualitative RGB representation of green, red, and blue representing the ball, offense, and defense, respectively. The historical images in Figure \ref{histmap} show all activity is near the hoop while the filter images in Figure \ref{filters1} show additional activity far away from the hoop. To compare Figures \ref{filters1} and \ref{histmap} quantitatively, we utilize the SSIM, a.k.a. the structural similarity index measure (\cite{wang2004image}). If a noisy version of an image is compared to the original, the SSIM can correctly identify that the images are the same.  SSIM ranges from $-1$ to $1$, where a perfect score of $1$ indicates that the two images are the same. 

We choose to compare the images from Figures \ref{filters1} and \ref{histmap} based upon our RGB representations. For example, previously we established that the red areas in image (1) of Figure \ref{filters1} are a representation of the offense. Therefore, we compare the offensive channel of image (1) in Figure \ref{filters1} to the historical offensive data of image (1) in Figure \ref{histmap}. We then compare image (2) of Figure \ref{filters1} to image (2) of Figure \ref{histmap} (historical ball data). We choose to compare image (3) of Figure \ref{filters1}, which we stated before represents both offensive and ball activity, to both the historical ball (2) and offense images (1) of Figure \ref{histmap}. After computing the SSIM for each part of the third filter of Figure \ref{filters1}, we average the two SSIM scores. Last, we compare image (4) of Figure \ref{filters1} to historical defensive data displayed in image (3) of Figure \ref{histmap}. We calculate the SSIM for two cases: the entire court and the half of the court containing the hoop. We choose these two cases because the filter images of Figure \ref{filters1} show activity away from the court while the images of Figure \ref{histmap} do not.  We detail the comparison images and resulting SSIM scores in Table \ref{ssim}.

%\bigskip

\begin{table}[h]
\captionof{table}{SSIM Results}
\centering
\begin{tabular}{llrr}
\cline{1-4}
Figure \ref{filters1} Image &Figure \ref{histmap} Image &SSIM (Half) &SSIM (Full) \\
\hline
Filter (1) &Offense (1) &0.623 &0.715 \\ 
Filter (2) &Ball (2) &0.310 &0.277 \\ 
Filter (3) &Offense, Ball (1,2) &0.554 & 0.580 \\ 
Filter (4) &Defense (3) &0.561 &0.680 \\
\lasthline
\end{tabular}
\label{ssim}
\end{table}

\bigskip

When calculating the SSIM for both half court and full court images, we expected that the full court image scores would be lower. However, as seen in Table \ref{ssim}, this is not the case for a majority of the images. Historical data from Figure \ref{histmap} shows that most activity is near the hoop; however, there is some activity far away from the hoop (likely due to transition plays). Since the filters of Figure \ref{filters1} capture this, the full court SSIM is larger for three out of four of the Figure \ref{filters1} images.

The worst performing image using SSIM as our evaluation tool is image (2) from Figure \ref{filters1} with an SSIM of $0.310$. This is due to a lack of large green areas near the hoop of the court in the filter image. When considering the entire court, image (2) from Figure \ref{filters1} shows ball activity in the corners away from the hoop. Since the historical ball image in Figure \ref{histmap} shows no activity in the upper corners, it is not surprising that it is the worst performing image.

Examining the results of Table \ref{ssim}, we can scrutinize the strength of our CNN. The accuracy of the CNN is close to that of the FFN, but the SSIM scores show room for improvement. It is clear that the CNN struggles to identify ball information when comparing image (2) of Figure \ref{filters1} to historical ball data.

\section{Conclusion}

Rather than other methods that incorporate transitions between a coarse and fine-grained approach or use small pieces of trajectory data, we utilize the full trajectory data of both offensive and defensive players. Since player alignments commonly permute, an image-based approach maintains the general spatial alignment of the players on the court. To integrate the time dependency of the trajectories, we introduce ``fading'' to our images to capture player paths. We found that using linear fading rather than a one-hot fade, works much better in our predictions. In addition, using a different color for each trajectory does not increase predictability in an RGB, three channel, image. Since traditional CNN's utilize three channels, we found that networks created to work with that kind of data, such as residual networks, AlexNet, and Network in Network underperformed. We therefore opt to build our own CNN to handle the trajectory images. Thus, by using our combined CNN and FFN, we can predict whether a shot is made with 61.5\% accuracy with the CNN proving to be more accurate than a FFN with hand-crafted features.

We found that by using a CNN, we can further explore the data using gradient ascent to picture the various filters of our network. We found that as the network gets deeper, it tends to gather several features together. For example, in the first layer, the filters look for shot locations. As we delve deeper into the network, the filters also begin to look for defensive and offensive spatial positions to make a more accurate prediction. Also, the histograms agree with common knowledge that centers have the highest short percentage since their shots tend to be right beside the basket. 

For further research, it would be very interesting to identify time dependency in basketball plays. In our image data, we subtract a flat amount at each equally spaced frame to cause the fading effect. However, this assumes that the data in time is linearly related. Since this is not necessarily true, designing a recurrent model to find this temporal dependency could be a very interesting problem. Instead of having a fading effect in the image data, we can design an LSTM that takes a moving window of player and ball trajectories.

One last aspect that was not considered during this study was the identities of teams and players. The focus of this research was to gather more insight on the average shooting plays of teams in the NBA. However, teams in the NBA have drastically different strategies. For example, the Golden State Warriors tend to rely on a three-point strategy while bigger teams, such the Thunder, build their offensive strategy around being inside the paint. Thus, new knowledge on basketball could be gathered if models were applied to different teams and possibly identify some overall team strategies. Such a more fine-grained analysis would require much more data.

\bibliographystyle{unsrt}
\bibliography{bballpaper}

\end{document}